\documentclass[final]{siamltex}
\usepackage{algorithm,algpseudocode}
\usepackage{latexsym, graphicx, epsfig, amsmath, amsfonts,amssymb,bm}
\usepackage{subcaption}
\usepackage{epstopdf}
\usepackage{booktabs}
\usepackage{array}
\usepackage{color}

\def\be{\begin{equation}}
\def\ee{\end{equation}}

\def\x{\mathbf{x}}
\def\y{\mathbf{y}}

\def\Nt{\widetilde{N}}
\def\Nh{\widehat{N}}
\def\Nhh{\widehat{\widehat{N}}}
\def\NN{\mathcal{N}}
\def\w{\mathbf{w}}

\def\wt{\widetilde{\w}}

\def\wh{\hat{\w}}

\def\bt{\widetilde{b}}

\def\S{\mathbb{S}}

\def\s{\sigma}
\def\Span{\textrm{Span}}

\def\E{{\mathbb E}}
\def\R{{\mathbb R}}

\title{Reducing Parameter Space for Neural Network Training}

\author{Tong Qin\and Ling Zhou \and
 Dongbin
       Xiu\thanks{Department of Mathematics,
		The Ohio State University, Columbus, OH 43210, USA.
		{\tt qin.428@osu.edu, zhou.2568@osu.edu, xiu.16@osu.edu.}
		}
}

\begin{document}
\maketitle

\begin{abstract}
For neural networks (NNs) with rectified linear unit (ReLU) or binary activation functions, we show that their training can be
accomplished in a reduced parameter space. Specifically, the weights in each neuron can be trained on the unit sphere, as opposed to the entire space, and the threshold can be trained in a bounded interval, as opposed to the real line.
We show that the NNs in the reduced parameter space are mathematically equivalent to the standard NNs with parameters in the whole space.
The reduced parameter space shall facilitate the optimization procedure for the network training, as the search space becomes
(much) smaller. We demonstrate the improved training performance using numerical  examples.
\end{abstract}


\section{Introduction} \label{sec:intro}

Interest in neural networks (NNs) has significantly increased in
recent years because of the successes of deep networks
in many practical applications.
Complex and
deep neural networks are known to be capable of learning very complex
phenomena that are beyond the capabilities of many other traditional
machine learning techniques. The amount of literature is too large to
mention. Here we cite just a few review type publications \cite{montufar2014number,
bianchini2014complexity, eldan2016power, poggio2017, DuSwamy2014,
GoodfellowBC-2016, Schmidhuber2015}. 

In a NN network, each neuron produces an
output in the following form
\begin{equation}
\s(\w\cdot \x_{in} + b),
\end{equation}
where the vector $\x_{in}$ represents the signal from all incoming
connecting neurons, $\w$ are the weights for the input, $\s$ is
the activation function, with $b$ as its threshold. In a complex (and
deep) network with a large number of neurons, the total number of the free
parameters $\w$ and $b$ can be exceedingly large.
Their training thus
poses a tremendous numerical challenge, 
as the objective function (loss function) to be optimized becomes
highly non-convex and with highly complicated landscape
(\cite{li2017visualizing}). Any numerical optimization procedures can
be trapped in a local minimum and produce unsatisfactory training results.
%

This paper is not concerned with numerical algorithm aspect of the
NN training. Instead,
the purpose of this paper is to show that the training of NNs can be
conducted in a reduced parameter space, thus providing any numerical
optimization algorithm a smaller space to search for the parameters.
This reduction applies to the type of activation
functions with the following scaling property: for any $\alpha\geq 0$,
$\s(\alpha\cdot y) = \gamma(\alpha) \s(y)$, where  $\gamma$ depends only on
$\alpha$. The binary activation function \cite{mcculloch1943logical}, one of the first used
activation functions, satisfies this property with $\gamma\equiv
1$. The rectified linear unit (ReLU) \cite{nair2010rectified, glorot2011deep}, one of the most widely used
activation functions nowadays, satisfies this property with $\gamma =
\alpha$. For NNs with this type of activation functions, we show that
they can be equivalently trained in a reduced parameter space. More
specifically, let the length of the weights $\w$ be $d$. Instead of training $\w$
in $\R^d$, they can be equivalently trained as unit vector with
$\|\w\|=1$, which means  $\w\in \S^{d-1}$, the unit sphere in $\R^d$. Moreover, if one is interested in approximating a function
defined in a bounded domain $D$, the
threshold can also be trained in a bounded interval $b\in [-X_B, X_B]$,
where $X_B=\sup_{\x\in D} \|\x\|$, as
opposed to the entire real line $b\in \R$. It is well known that the
standard NN with single hidden layer is a universal approximator, c.f. \cite{pinkus1999,
  barron1993universal, hornik1991approximation}. Here we prove that
our new formulation in the reduced parameter space is also a universal
approximator, in the sense that its span is dense in $C(\R^d)$. We
then further extend the parameter space constraints to NNs with
multiple hidden layers. The major advantage of the constraints in the
weights and thresholds is that they significantly reduce the search
space for these parameters during training. Consequently, this eliminates a potentially
large number of undesirable local minima that may cause training
optimization algorithm to terminate prematurely, which is one of the
reasons for unsatisfactory training results. We then present
 examples for function approximation to verify this numerically. Using
 the same network structure, optimization solver, and identical random
 initialization, our numerical tests show that the training results in
 the new reduced parameter space is notably better than those from the
 standard network. More importantly, the training using the reduced
 parameter space is much more robust against initialization.

This paper is organized as follows. In Section \ref{sec:single}, we first
derive the constraints on the parameters using network with single
hidden layer, while enforcing the equivalence of the
network. In
Section \ref{sec:proof}, we prove that the constrained NN formulation remains a
universal approximator. In Section \ref{sec:multiple}, we present the constraints for
NNs with multiple hidden layers. Finally in Section \ref{sec:examples}, we present
numerical experiments to demonstrate the improvement of the network
training using the reduced parameter space. We emphasize that this
paper is not concerned with any particular training
algorithm. Therefore, in our numerical tests we used the most standard
optimization algorithm from Matlab\textregistered

\section{Constraints for NN with Single Hidden Layer} \label{sec:single}
Let us first consider the standard NN with a single hidden layer, in the
context for approximating an unknown response function $f:\R^d \to
\R$. 
The NN approximation using activation function $\s$ takes the following form,
\begin{equation}  \label{Nx}
	N(\x) = \sum_{j=1}^N c_j\s(\w_j\cdot\x + b_j),\quad \x\in\R^d,
\end{equation}
where $\w_j\in\R^d$ is the weight vector, $b_j\in\R$ the threshold,
$c_j\in\R$,  and $N$ is the width of the network.


We restrict our discussion to following two activation functions. 
One is the rectified linear unit (ReLU),
\begin{equation} 
	\label{ReLU}
	\s(x) = \max(0, x).
\end{equation}

The other one is the binary activation function, also known as heaviside/step function,
\begin{equation}  
	\label{step}
	\s(x) = \left\{
	\begin{array}{ll}
	1, & x> 0,\\
	0, & x<0,
	\end{array}
	\right.
\end{equation}
with $\s(0)=\frac12$.

We remark that these two activation functions satisfy the following
scaling property:
For any $y\in \R$ and $\alpha\geq 0$, there exists a constant
$\gamma\geq 0$ such that
\begin{equation}   \label{scale}
\s (\alpha\cdot y) = \gamma(\alpha)\s(y), 
\end{equation}
where $\gamma$ depends only on $\alpha$ but not on $y$.
The ReLU satisfies this property with $\gamma(\alpha) = \alpha$, which
is known as {\em scale invariance}. The binary activation function
satisfies the scaling property with $\gamma(\alpha)\equiv 1$. 

We also list the following properties, which are important for the
method we present in this paper.
		\begin{itemize}
			\item For the binary activation function
                          \eqref{step},  for any $x\in\R$,
				\begin{equation} 
					\label{step-1}
					\s(x) +\s(-x)\equiv 1.
				\end{equation}
			\item For the ReLU activation function
                          \eqref{ReLU}, for any $x\in\R$ and any $\alpha$,
				\begin{equation} 
					\label{ReLU-1}
					L(x;\alpha):=\s\left(x+\frac{\alpha}{2}\right) -\s\left(x-\frac{\alpha}{2}\right) +
					\s\left(-x+\frac{\alpha}{2}\right) - \s\left(-x-\frac{\alpha}{2}\right)\equiv\alpha.
				\end{equation}
		\end{itemize}


\subsection{Weight Constraints}

We first show that the training of \eqref{Nx} can be equivalently conducted
with constraint $\|\w_j\|=1$, i.e., unit vector. This is a
straightforward result of the scaling property \eqref{scale} of the
activation function. It effectively
reduces the search space for the weights from $\w_j\in\R^d$ to $\w_j\in\S^{d-1}$, the
unit sphere in $\R^d$.

\begin{proposition}
	\label{prop:eq1}
Any neural network construction \eqref{Nx} using the ReLU \eqref{ReLU}  or the binary \eqref{step}  activation
functions has an
equivalent form
\begin{equation}  \label{Nnew}
\Nt(\x) = \sum_{j=1}^{\Nt} \widetilde{c}_j\s(\wt_j\cdot\x + \widetilde{b}_j), \quad \|\wt_j\|=1.
\end{equation}
\end{proposition}

\begin{proof}
Let us first assume $\|\w_j\|\neq 0$ for all $j=1,\dots,N$. We then have
\begin{equation*}  
\begin{split}
N(\x) &= \sum_{j=1}^N c_j\s(\w_j\cdot\x + b_j) \\
         & =\sum_{j=1}^N c_j
         \s\left(\|\w_j\|\left(\frac{\w_j}{\|\w_j\|}\cdot\x +
             \frac{b_j}{\|\w_j\|}\right)\right) \\
&= \sum_{j=1}^N c_j \gamma(\|\w_j\|) 
\s\left(\frac{\w_j}{\|\w_j\|}\cdot\x +
            \frac{b_j}{\|\w_j\|}\right),
\end{split}
\end{equation*}
where $\gamma$ is the factor in the scaling property \eqref{scale}
satisfied by both ReLU and binary activation functions. Upon defining
\begin{equation*}
	\widetilde{c}_j = c_j \gamma(\|\w_j\|), \quad
	\wt_j = \frac{\w_j}{\|\w_j\|}, \quad
	\bt_j = \frac{b_j}{\|\w_j\|},
\end{equation*}
we have the following equivalent form as in \eqref{Nnew}
\begin{equation*}
	N(\x)=\sum_{j=1}^N \widetilde{c}_j\s(\wt_j\cdot\x+\bt_j).
\end{equation*}

Next, let us consider the case $\|\w_j\|=0$ for some
$j\in\{1,\dots,N\}$. The contribution of this term to the construction
\eqref{Nx} is thus
\begin{equation*}
c_j \s(b_j) =\left\{
\begin{array}{ll}
\widehat{c}_j, & b_j\geq 0,\\
0, & b_j<0,
\end{array}
\right.
\end{equation*} 
where $\widehat{c}_j = c_j$ for the binary activation function and $\widehat{c}_j = c_j
b_j$ for the ReLU function. Therefore, if $b_j<0$, this term in
\eqref{Nx} vanishes. If $b_j\geq 0$, the contributions of these terms
in \eqref{Nx} are constants, which can be represented by a combination
of neuron outputs using the relations \eqref{step-1} and \eqref{ReLU-1}, for
binary and ReLU activation functions, respectively. 
We thus obtain a new representation in the form of \eqref{Nnew} that
includes all the terms in the original expression \eqref{Nx}.
This completes the proof.
\end{proof}

The proof immediately gives us another equivalent form, by combining
all the constant terms from \eqref{Nx} into a single constant first
and then explicitly including it in the expression.
\begin{corollary}
Any neural network construction \eqref{Nx} using the ReLU \eqref{ReLU}  or the binary \eqref{step}  activation
functions has an
equivalent form
\begin{equation}  \label{Nnew0}
	\Nt(\x) = \widetilde{c}_0 + \sum_{j=1}^{\tilde{N}} \widetilde{c}_j\s(\wt_j\cdot\x + \widetilde{b}_j), \quad \|\wt_j\|=1.
\end{equation}.
\end{corollary}


\subsection{Threshold Constraints}

We now present constraints on the thresholds in \eqref{Nx}. This is
applicable when
the target function $f$ is defined on a bounded domain, i.e.,
$f:D\to \R$, with $D\subset \R^d$ bounded. This is often the case
in practice. We demonstrate that any NN \eqref{Nx} can be trained
equivalently in a bounded interval for each of the thresholds. This (significantly)
reduces the search space for the thresholds.

\begin{proposition}
	\label{prop:eq2}
With the ReLU \eqref{ReLU}  or the binary \eqref{step}  activation
function,
let \eqref{Nx} be an approximator to a function $f: D\to \R$, where
$D\subset \R^d$ is a bounded domain. Let 
\begin{equation} \label{Xb}
X_B = \sup_{\x\in D} \|\x\|.
\end{equation}
Then, \eqref{Nx} has an equivalent form
\begin{equation}  \label{Nnew2}
\Nh(\x) = \sum_{j=1}^{\Nh} \hat{c}_j\s(\wh_j\cdot\x +
\hat{b}_j), \quad \|\wh_j\|=1, \quad -X_B\leq \hat{b}_j
\leq X_B.
\end{equation}
\end{proposition}
\begin{proof}
	Proposition \ref{prop:eq1} establishes that \eqref{Nx} has an
	equivalent form \eqref{Nnew}
	\begin{equation*}  
	\Nt(\x) = \sum_{j=1}^{\Nt} \widetilde{c}_j\s(\wt_j\cdot\x + \widetilde{b}_j), \quad \|\wt_j\|=1.
	\end{equation*}
	Since $\wt_j$ is a unit vector, we have
	\begin{equation} 
	\wt_j\cdot \x \in \left[-\|\x\|,\|\x\|\right] \subseteq [-X_B, X_B],
	\end{equation}
	where the bound \eqref{Xb} is used.

	Let us first consider the case $\widetilde{b}_j < -X_B$, then $\forall
	\x\in D$, $\wt_j\cdot \x
	+ \widetilde{b}_j <0$, $\s(\wt_j\cdot \x
	+ \widetilde{b}_j) = 0$, for both the ReLU and binary activation
	functions. Subsequently, this term has no contribution to the
	approximation and can be eliminated.

	Next, let us consider the case $\widetilde{b}_j > X_B$, then $\wt_j\cdot \x
	+ \widetilde{b}_j >0$ for all $\x\in D$.
	Let $J=\{j_1,\dots j_L\}$, $L\geq 1$, be the set of terms in
	\eqref{Nnew} that satisfy this condition.
	We then have
	$\wt_{j_\ell}\cdot \x + \widetilde{b}_{j_\ell} >0$, for all $\x\in D$, and
	$\ell=1,\dots, L$. We now show that the net contribution of these terms
	in \eqref{Nnew} is included in the equivalent form \eqref{Nnew2}.
\begin{enumerate}
\item
	For the binary activation function \eqref{step}, the contribution of
	these terms to the approximation \eqref{Nnew} is
	\begin{equation*} 
	N_J(\x) = \sum_{\ell=1}^L \widetilde{c}_{j_\ell} \s\left(\wt_{j_\ell}\cdot \x +
	  \widetilde{b}_{j_\ell}\right) 
	= \sum_{\ell=1}^L \widetilde{c}_{j_\ell} = \const.
	\end{equation*}
	Again, using the relation \eqref{step-1}, any constant can be 
	 expressed by a combination of binary activation terms with
         thresholds $\widetilde{b}_j = 0$.
Such terms are already included in \eqref{Nnew2}. 

\item
	For the ReLU activation \eqref{ReLU}, the contribution of these terms
	to \eqref{Nnew} is
	\begin{equation} 
	\begin{split}
	N_J(\x) &= \sum_{\ell=1}^L \widetilde{c}_{j_\ell} \s\left(\wt_{j_\ell}\cdot \x +
	  \widetilde{b}_{j_\ell}\right) \\
	&= \sum_{\ell=1}^L \widetilde{c}_{j_\ell} \left(\wt_{j_\ell}\cdot \x +
	  \widetilde{b}_{j_\ell}\right) \\
	&= \sum_{\ell=1}^L \left(\widetilde{c}_{j_\ell} \wt_{j_\ell}\right)\cdot \x +
	  \sum_{\ell=1}^L \widetilde{c}_{j_\ell} \widetilde{b}_{j_\ell} \\
	&= \wt^*_J \cdot \x + b^*_J \\
	&= \s(\wt^*_J \cdot \x) - \s(-\wt^*_J \cdot \x)  + b^*_J,
	\end{split}
	\end{equation}
where the last equality follows the simple property of the ReLU
function $\s(y) - \s(-y) = y$.
	Using Proposition \ref{prop:eq1}, the first two terms then have an equivalent
	form using unit weight $\wt_J^* = \wt_J^*/\|\wt_J^*\|$ and with
	zero threshold, which is included in \eqref{Nnew2}. For the
	constant $b_J^*$, we again invoke the relation \eqref{ReLU-1}
        and represent it by $\frac{b_J^*}{\alpha} L(\wh_J^*\cdot\x,
        \alpha)$, where $\wh_J^*$ is an arbitrary unit vector and $0<\alpha<X_B$.
	 Obviously, this expression includes a collection of terms (4
         terms as in \eqref{ReLU-1}), which are 
         included in \eqref{Nnew2}. This completes the proof.
\end{enumerate}
\end{proof}


The equivalence between the standard NN expression \eqref{Nx} and the
constrained expression \eqref{Nnew2} indicates that the NN training
can be conducted in a reduced parameter space. For the weights $\w_j$ in each
neuron, its training can be conducted in $\S^{d-1}$, the
$d$-dimensional unit sphere, as opposed to the entire space
$\R^d$. For the threshold, its training can be conducted in the
bounded interval $[-X_B, X_B]$, as opposed to the entire real line
$\R$. The reduction of the parameter space can eliminate many
potential local minima and therefore enhance the performance of numerical
optimization during the training. We remark that the equivalent form
in the reduced parameter space \eqref{Nnew2} has different number of ``active''
neurons than the original unrestricted case \eqref{Nx}.

\section{Universal approximation property} \label{sec:proof}

By universal approximation property, we aim to establish that the constrained
formulations \eqref{Nnew} and \eqref{Nnew2} can approximate any continuous
functions. To this end, we define the following set of functions on $\R^d$
\begin{equation}
	\NN (\s;\Lambda,\Theta):=\Span\left\{\s(\w\cdot\x+b): \w\in \Lambda,b\in \Theta \right\},
\end{equation}
where $\Lambda\in \R^d$ is the weight set, $\Theta\in\R$ the threshold
set.  We also denote $\NN_D(\s; \Lambda, \Theta)$ as the same set of functions when confined in a bounded domain $D\subseteq \R^d$. 

By following this definition, the standard NN expression and our two
constrained expressions correspond to the following spaces.
\begin{equation}
\begin{split}
\eqref{Nx} &\in  \NN(\s; \R^d, \R), \\
\eqref{Nnew} &\in \NN(\s; \S^{d-1}, \R), \\
\eqref{Nnew2}  &\in \NN_D(\s; \S^{d-1}, [-X_B, X_B]),
\end{split}
\end{equation}
where $\S^{d-1}$ is the unit sphere in $\R^d$ because
$\|\wt\|=1$. 

The universal approximation property for the standard unconstrained NN
expression \eqref{Nx} has
been well studied, cf. \cite{cybenko1989approximation,
hornik1991approximation, mhaskar1992approximation,
barron1993universal, leshno1993multilayer}, and \cite{pinkus1999} for a
survey.
Here we cite the following result for $\NN(\s; \R^d, \R)$.
\begin{theorem}[\cite{leshno1993multilayer}, Theorem 1]
	\label{propLeshno}
	Let $\s$ be a function in $L_{loc}^\infty(\R)$, of which the set of
	discontinuities has Lebesgue measure zero. Then the set $\NN(\s; \R^d, \R)$
	is dense in $C(\R^d)$, in the topology of uniform convergence on compact sets, 
	if and only if $\s$ is not an algebraic polynomial
	almost everywhere.
\end{theorem}

\subsection{Universal approximation property of \eqref{Nnew}}
\label{sec:Dense1} 

We now examine the universal approximation property for the first
constrained formulation \eqref{Nnew}.

\begin{theorem}
	\label{thm1}
	Let $\s$ be the binary function \eqref{step} or the ReLU function
	\eqref{ReLU}, then we have 
	\begin{equation}
		\label{thm1eq}
		\NN(\s; \S^{d-1}, \R)=\NN(\s; \R^d, \R).
	\end{equation}	
 	and the set $ \NN(\s; \S^{d-1}, \R)$ is dense in $C(\R^d)$, in the topology of uniform convergence on compact sets.
\end{theorem}

\begin{proof}
	Obviously, we have $\NN(\s; \S^{d-1}, \R)\subseteq\NN(\s; \R^d, \R)$. By
	Proposition \ref{prop:eq1}, any element $N(\x)\in \NN(\s; \R^d, \R) $ can
	be reformulated as an element $\Nt(\x)\in \NN(\s; \S^{d-1}, \R)$.
	Therefore, we have $\NN(\s; \R^d, \R)\subseteq \NN(\s; \S^{d-1}, \R)$. This
	concludes the first statement \eqref{thm1eq}. Given the
        equivalence \eqref{thm1eq}, the denseness result immediately
        follows from Theorem \ref{propLeshno}, as both the ReLU and the
        binary activation functions are not polynomials and are
        continuous everywhere except at a set of zero Lebesgue measure.
\end{proof}

\subsection{Universal approximation property of \eqref{Nnew2}}

We now examine the second constrained NN expression \eqref{Nnew2}.
\begin{theorem}
	Let $\s$ be the binary \eqref{step} or the ReLU \eqref{ReLU} activation
	function. Let $\x\in D\subset \R^d$, where $D$ is closed and bounded with
        $X_B=\sup_{\x\in D} \|\x\|$. Define $\Theta=[-X_B, X_B]$, then
	\begin{equation}
		\label{thm2eq}
		\NN_D(\s; \S^{d-1}, \Theta)=\NN_D(\s;\R^{d}, \R).
	\end{equation}
	Furthermore, $\NN_D(\s; \S^{d-1}, \Theta)$ is dense in
        $C(D)$ in the topology of uniform convergence.
\end{theorem}
\begin{proof}
	Obviously we have $\NN_D(\s; \S^{d-1},
	\Theta)\subseteq \NN_D(\s; \R^d, \R)$. On the other hand,
        Proposition \ref{prop:eq2} establishes that for any element $N(\x)\in
	\NN_D(\s; \R^{d}, \R)$, there exists an
	equivalent formulation $\Nh(\x)\in \NN_D(\s; \S^{d-1}, \Theta)$
        for any $\x\in D$. This implies $\NN_D(\s;\R^d,
        \R)\subseteq \NN_D(\s; \S^{d-1}, \Theta)$. We then have
	\eqref{thm2eq}. 

	For the denseness of $\NN_D(\s; \S^{d-1}, \Theta)$ in
	$C(D)$, let us consider any function $f\in C(D)$. By the Tietze extension
	theorem (cf. \cite{folland1999}), there exists an extension $F\in C(\R^d)$ with
	$F(\x)=f(\x)$ for any $\x\in D$. Then, the denseness
        result of the standard unconstrained NN expression (Theorem
        \ref{propLeshno}) implies that, for the compact set $E=\{\x\in \R^d: \|\x\|\leq X_B\}$ and any given $\epsilon >0$,
        there exists $N(\x)\in \NN(\s; \R^{d}, \R)$ such that
	\begin{equation*}
		\sup_{\x\in\E} |N(\x)-F(\x)|\leq\epsilon.
	\end{equation*}
       By Proposition \ref{prop:eq2}, there exists an equivalent
       constrained NN expression $\Nh(\x)\in
	\NN_D(\s; \S^{d-1}, \Theta)$ such that $\Nh(\x)=N(\x)$ for any $\x\in
	D$. We then immediately have, for any
$f\in C(D)$ and any given $\epsilon>0$, there exists
$\Nh(\x)\in \NN_D(\s; \S^{d-1}, \Theta)$ such that  
	\begin{equation*}
		\sup_{\x\in D} |\Nh(\x)-f(\x)|=\sup_{\x\in D}|N(\x)-f(\x)|\leq \sup_{\x\in E}|N(\x)-F(\x)|\leq \epsilon.
	\end{equation*}
	The proof is now complete.
\end{proof}

\section{Constraints for NNs with Multiple Hidden
  Layers} \label{sec:multiple}

We now generalize the previous result to feedforward NNs with multiple
hidden layers. Let us again consider approximation of  
a multivariate function $f:D\to\R$, where $D\subseteq \R^d$ with
$\sup_{\x\in D}\|\x\|=X_B<\infty$. 

Consider a feedforward NN with $M$ layers, $M\geq 3$, where $m=1$ is
the input layer and $m=M$ the output layer. Let $J_m$, $m=1,\dots,M$
be the number of neurons in each layer. Obviously, we have $J_1=d$ and $J_M=1$ in our case. 
Let $\y^{(m)}\in \R^{J_m}$ be the output of the neurons in the $m$-th layer. Then, by following
the notation from \cite{DuSwamy2014}, we can write
\begin{equation}\label{MultNx}
\begin{split}
\y^{(1)} &= \x, \\
\y^{(m)} &= \s\left(\left[\mathbf{W}^{(m-1)}\right]^T \y ^{(m-1)}+\mathbf{b}^{(m)}\right), \qquad m=2,\dots, M-1,\\
\y^{(M)} &= \left[\mathbf{W}^{(M-1)}\right]^T \y^{(M-1)}.
\end{split}
\end{equation} 
where $\mathbf{W}^{(m-1)}\in \R^{J_{m-1}\times J_m}$ is the weight
matrix and $\mathbf{b}^{(m)}$ is the threshold vector.
In component form, the
output of the $j$-th neuron in the $m$-th layer is 
\begin{equation} \label{NN_layer}
y_j^{(m)} = \s\left( \left[\w_j^{(m-1)}\right]^T \y^{(m-1)} + b_j^{(m)}\right), \qquad j=1,\dots, J_m, \quad m=2,\dots, M-1,
\end{equation}
where $\w^{(m-1)}_{j}$ be the $j$-th column of $\mathbf{W}^{(m-1)}$.

\subsection{Weight constraints}

The derivation for the constraints on the weights vector
$\w_j^{(m-1)}$ can be generalized directly from the single-layer case
and we have the following weight constraints, 
\begin{equation} \label{weight_general}
\left\|\w^{(m-1)}_{j}\right\| = 1, \qquad j=1,\dots, J_m, \quad m=2,\dots, M-1.
\end{equation}

\subsection{Threshold constraints}

The constraints on the threshold $b_j^{(m)}$ depend on the bounds of the output from the previous layer $\y^{(m-1)}$.

For the ReLU activation function \eqref{ReLU}, we derive from \eqref{NN_layer} that for $m=2,\dots,M$,
\begin{equation} 
\begin{split}
\left|y_j^{(m)}\right| &\leq \left| \left[\w_j^{(m-1)}\right]^T \y^{(m-1)} + b_j^{(m)}\right| \\
&\leq  \left\|\w_j^{(m-1)}\right\| \left\|\y^{(m-1)}\right\| + \left|b_j^{(m)}\right| \\
&\leq  \left\|\y^{(m-1)}\right\| + \left|b_j^{(m)}\right|
\end{split}
\end{equation}

If the domain $D$ is bounded and with $X_B = \sup_{\x\in D} \|\x\|
<+\infty$, then the constraints on the threshold can be recursively
derived. Starting from $\|\y^{(1)}\| = \|\x\|\in [-X_B, X_B]$ and
$b_j^{(2)}\in [-X_B,X_B]$, we then have
\begin{equation} \label{b_general}
	b_j^{(m)} \in X_B\cdot[-2^{m-2}, 2^{m-2}], \qquad m=2,\dots,M-1, \quad j=1,\dots, J_m.
\end{equation}

For the binary activation function \eqref{step}, we derive from \eqref{NN_layer} that for $m=2,\dots,M-1$,
\begin{equation} 
\left|y_j^{(m)}\right| \leq 1.
\end{equation}
Then, the bounds for the thresholds are
\begin{equation}
\begin{split}
b_j^{(2)} &\in X_B\cdot[-1, 1],  \qquad j=1,\dots, J_2,\\
b_j^{(m)} &\in [-1, 1], \qquad m=3,\dots,M-1, \quad j=1,\dots, J_m.
\end{split}
\end{equation}

\section{Numerical Examples} \label{sec:examples}

In this section we present numerical examples to demonstrate the
properties of the constrained NN training. We focus exclusively on the
ReLU activation function \eqref{ReLU} due to its overwhelming
popularity in practice.

Given a set of training samples $\{(\x_i,
y_i)\}_{i=1}^n$, the weights and thresholds are trained by minimizing the
following mean squared error (MSE)
\begin{equation*}
	E=\frac{1}{n}\sum_{i=1}^n |N(\x_i)-y_i|^2.
\end{equation*}
We conduct the training using the standard unconstrained NN formulation \eqref{Nx}
and our new constrained formulation
\eqref{Nnew2} and compare the training results.
In all tests, both formulations use exactly the same randomized
initial conditions for the weights and thresholds.
Since our new constrained formulation is irrespective of
the numerical optimization algorithm, we use one of the most 
accessible optimization routines from MATLAB\textregistered,   
the function \texttt{fminunc} for unconstrained optimization and the function
\texttt{fmincon} for constrained optimization. It is natural to
explore the specific form of the constraints in \eqref{Nnew2} to design
more effective constrained optimization algorithms. This is, however,
out of the scope of the current paper.

After training the networks, we evaluate the network approximation errors
using another set of samples --- a validation sample set, which
consists of randomly generated points that are
independent of the training sample set. 
Even though our discussion applies to functions in arbitrary dimension
$d\geq 1$, we present only the numerical results in $d=1$ and $d=2$
because they can be easily visualized.

\subsection{Single Hidden Layer}

We first examine the approximation results using NNs with single
hidden layer, with and without constraints.

\subsubsection{One dimensional tests}\label{example1}

We first consider a one-dimensional smooth function 
\begin{equation}\label{testf1}
	f(x)=\sin(4\pi x), \quad x\in [0, 1].
\end{equation}
The constrained formulation \eqref{Nnew2} becomes
\begin{equation}  \label{Nx_1D1}
\Nh(\x) = \sum_{j=1}^{\Nh} \hat{c}_j\s(\hat{w}_j x +
\hat{b}_j), \quad \hat{w}_j\in \{-1,1\}, \quad -1 \leq \hat{b}_j
\leq 1.
\end{equation}
Due to the simple form of the weights and the domain $D=[0,1]$,
the proof of Proposition \ref{prop:eq2} also gives us the following
tighter bounds for the thresholds for this specific problem,
\begin{equation} \label{Nx_1D2}
\left\{
\begin{split}
-1\leq \hat{b}_j \leq 0,  &\qquad \textrm{if } \hat{w}_j = 1;\\
0\leq \hat{b}_j \leq 1, &\qquad \textrm{if } \hat{w}_j=-1.
\end{split}
\right.
\end{equation}

\begin{figure}[ht] 
  \begin{subfigure}[b]{0.5\linewidth}
	\label{fig1:con0}
    \centering
	\includegraphics[width=1\linewidth]{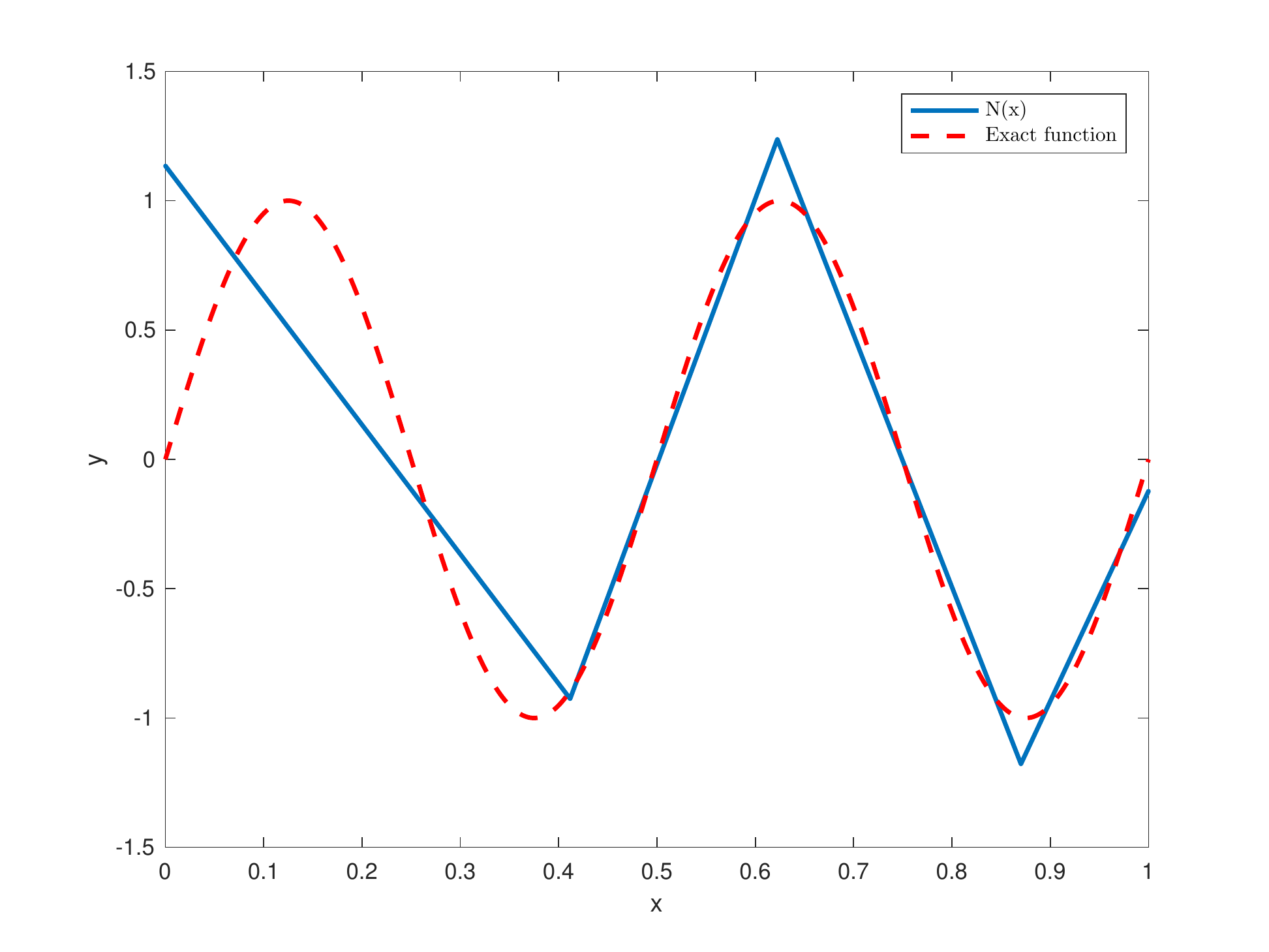} 
	\caption{Unconstrained NN $N(x)$ \eqref{Nx}} 
  \end{subfigure}
  \begin{subfigure}[b]{0.5\linewidth}
	\label{fig1:con1}
    \centering
	\includegraphics[width=1\linewidth]{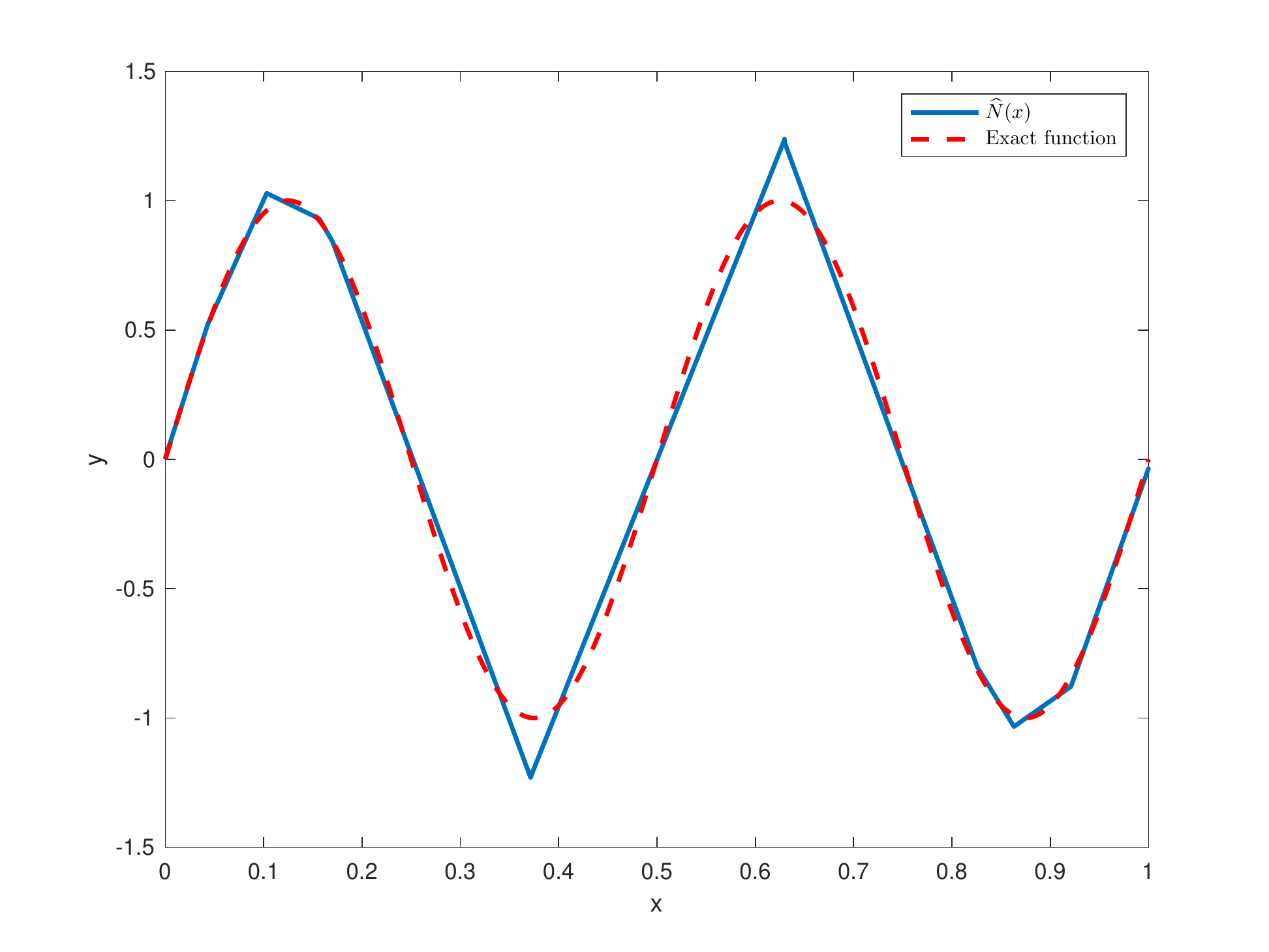} 
	\caption{Constrained NN $\Nh(x)$ \eqref{Nx_1D1}} 
  \end{subfigure} 
  \begin{subfigure}[b]{0.5\linewidth}
	\label{fig1:con2}
    \centering
	\includegraphics[width=1\linewidth]{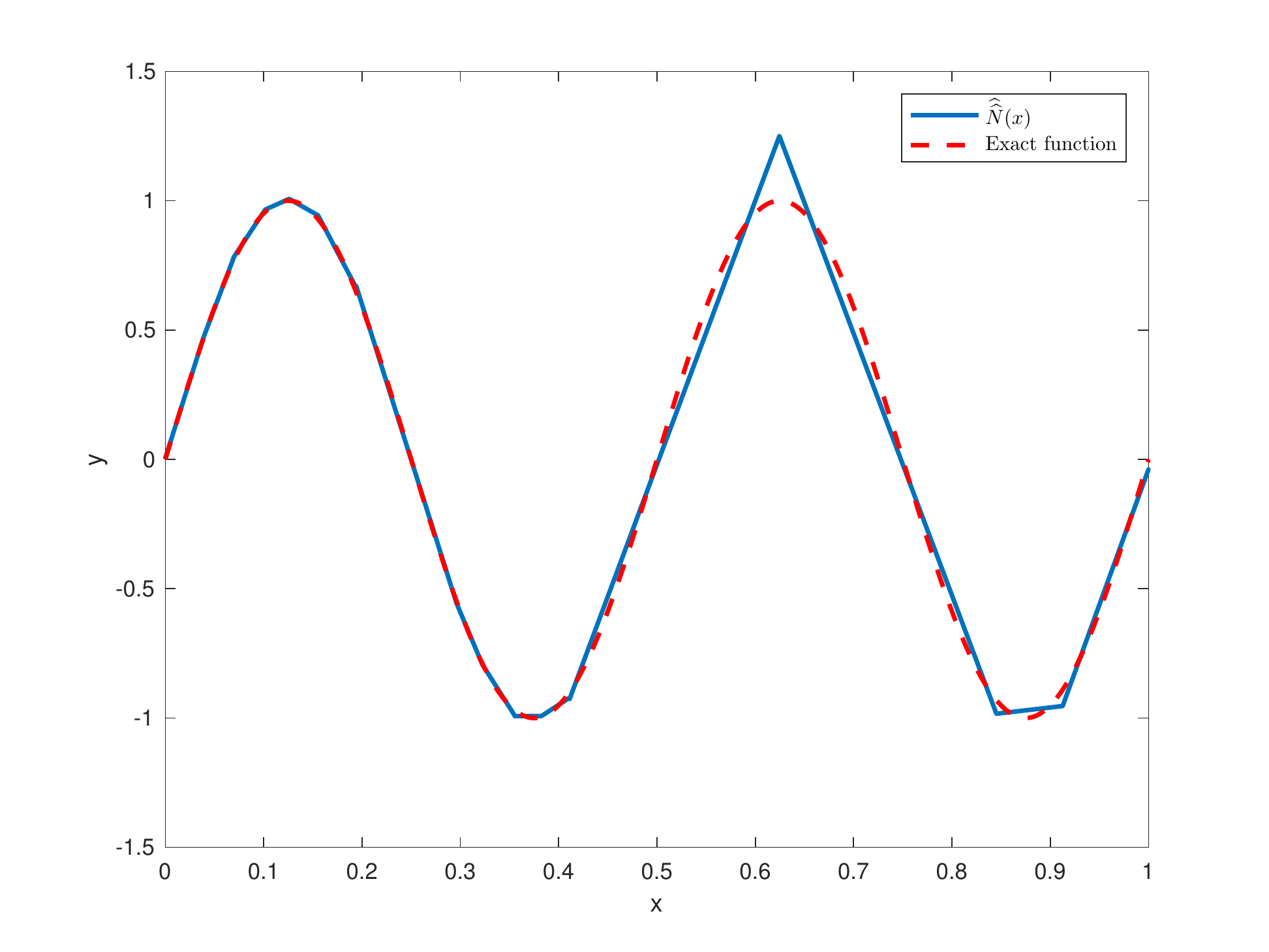} 
	\caption{Constrained NN $\Nhh(x)$ \eqref{Nx_1D2}}
  \end{subfigure}
  \begin{subfigure}[b]{0.5\linewidth}
	\label{fig1:MSE}
    \centering
	\includegraphics[width=1\linewidth]{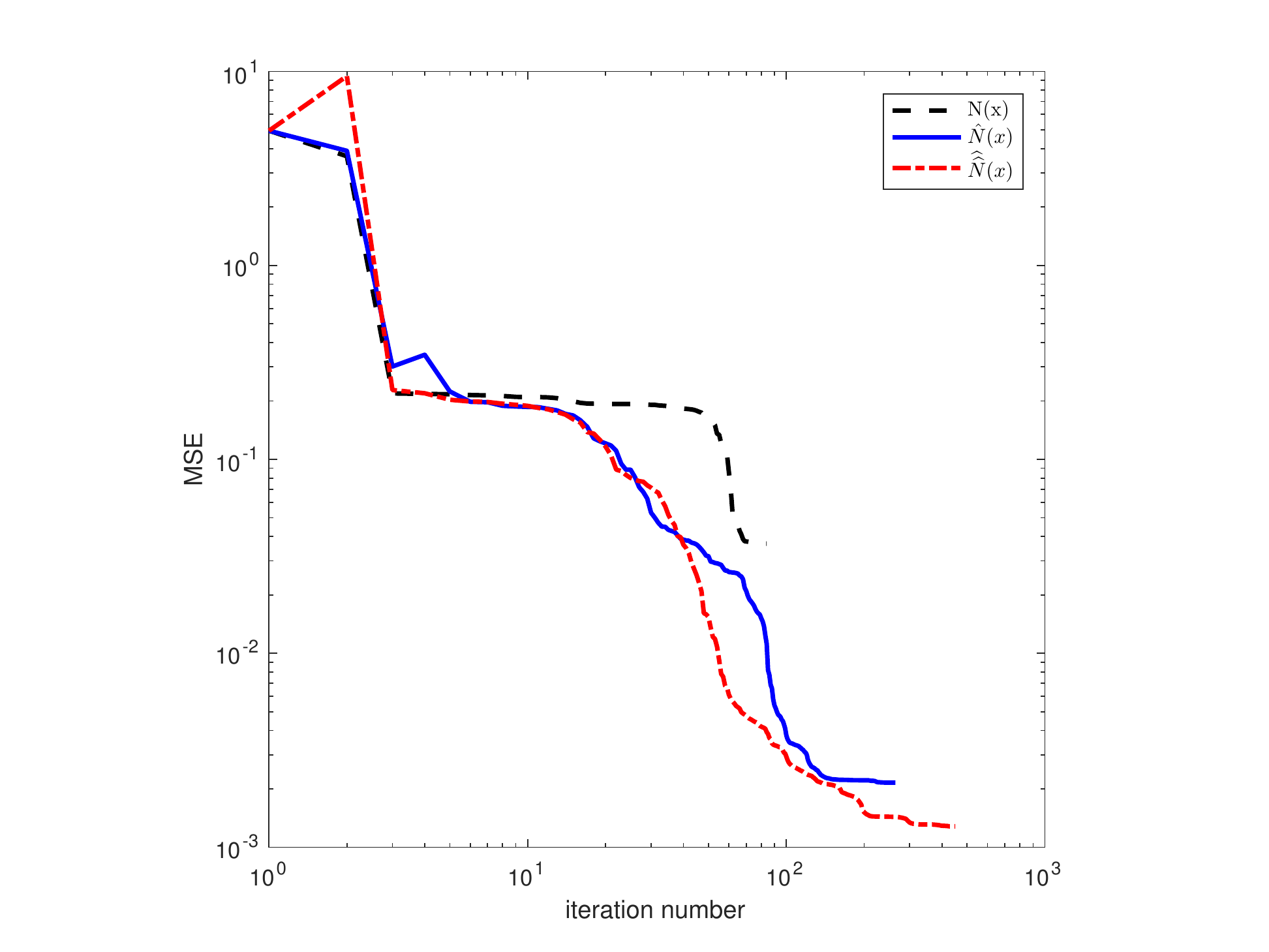} 
    \caption{Convergence history} 
  \end{subfigure} 
  \caption{Numerical results for \eqref{testf1} with one
    sequence of 
    random initialization.}
\label{fig:1L1D_IC1}
\end{figure}

\begin{figure}[ht] 
  \begin{subfigure}[b]{0.5\linewidth}
    \centering
	\includegraphics[width=1\linewidth]{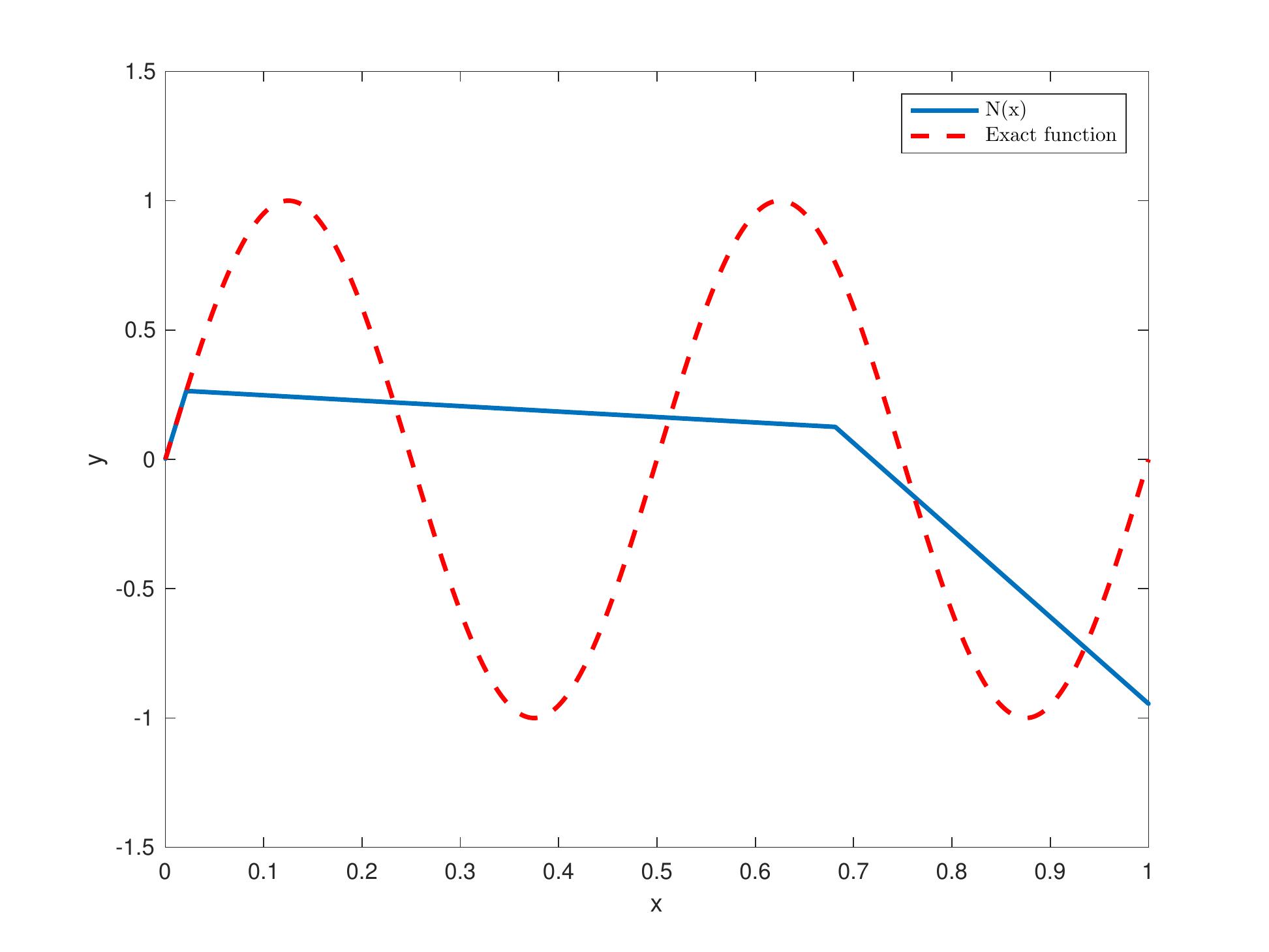} 
	\caption{Unconstrained NN $N(x)$ \eqref{Nx}} 
  \end{subfigure}
  \begin{subfigure}[b]{0.5\linewidth}
    \centering
	\includegraphics[width=1\linewidth]{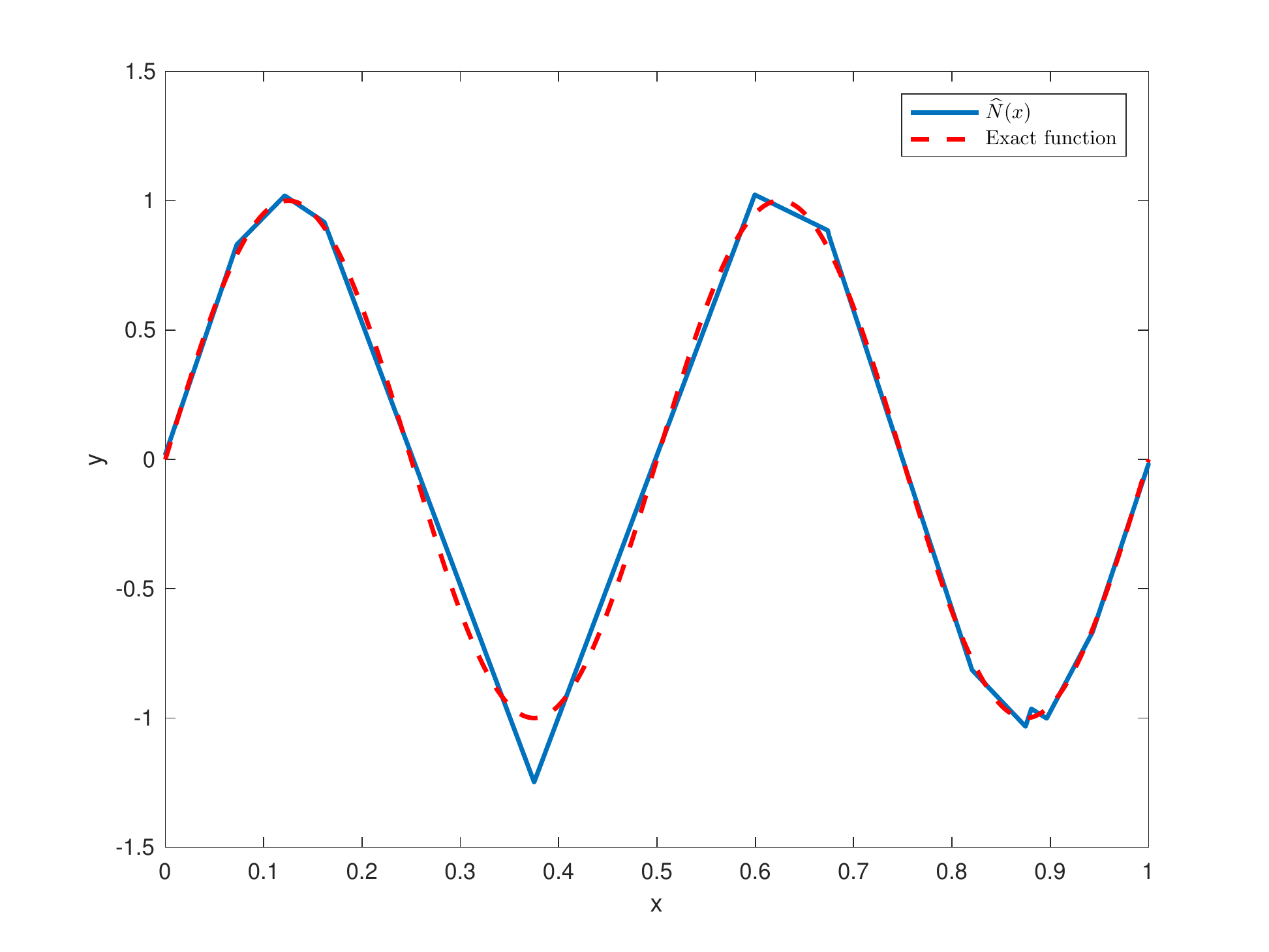} 
	\caption{Constrained NN $\Nh(x)$ \eqref{Nx_1D1}} 
  \end{subfigure} 
  \begin{subfigure}[b]{0.5\linewidth}
    \centering
	\includegraphics[width=1\linewidth]{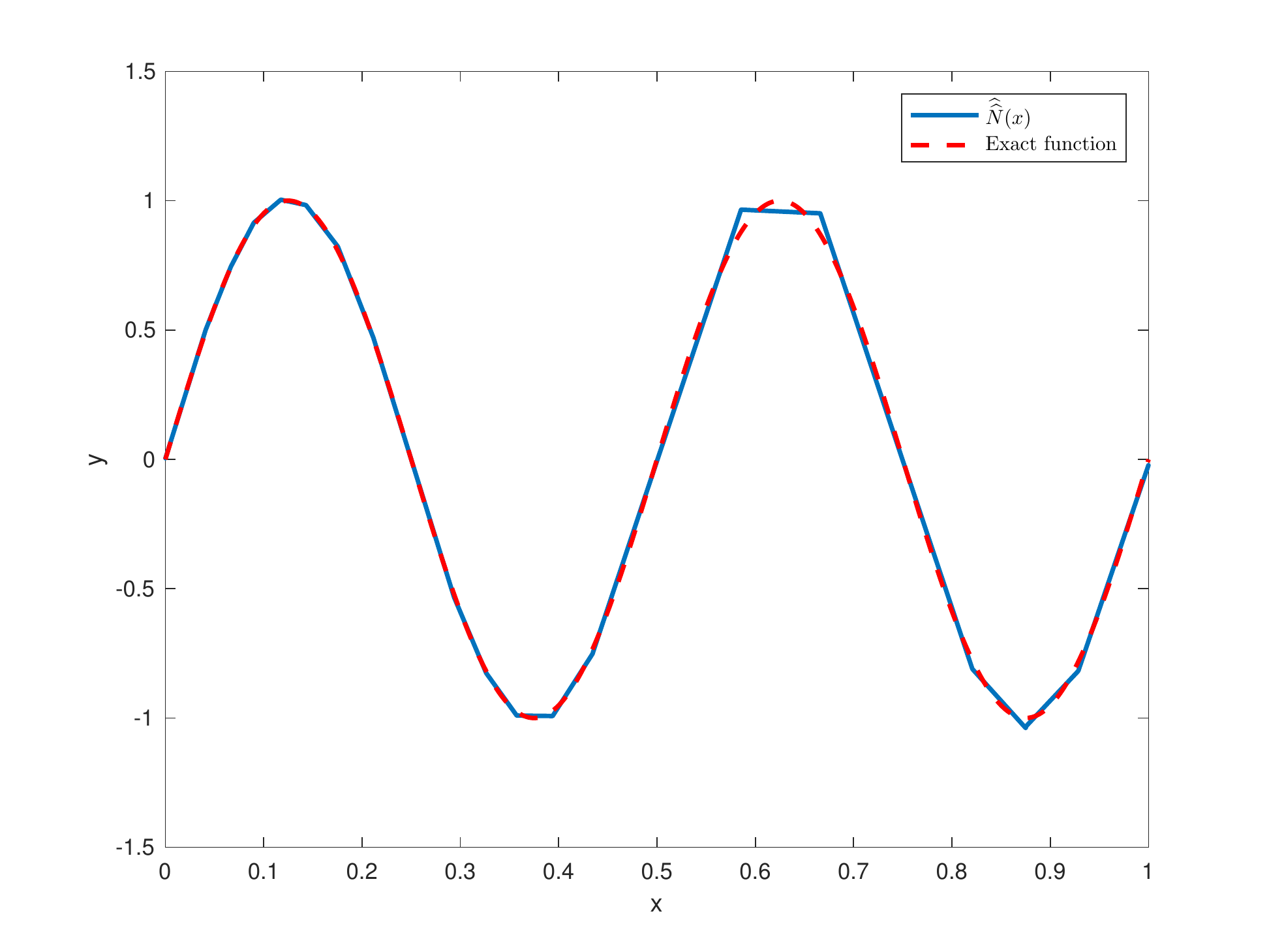} 
	\caption{Constrained NN $\Nhh(x)$ \eqref{Nx_1D2}}
  \end{subfigure}
  \begin{subfigure}[b]{0.5\linewidth}
    \centering
	\includegraphics[width=1\linewidth]{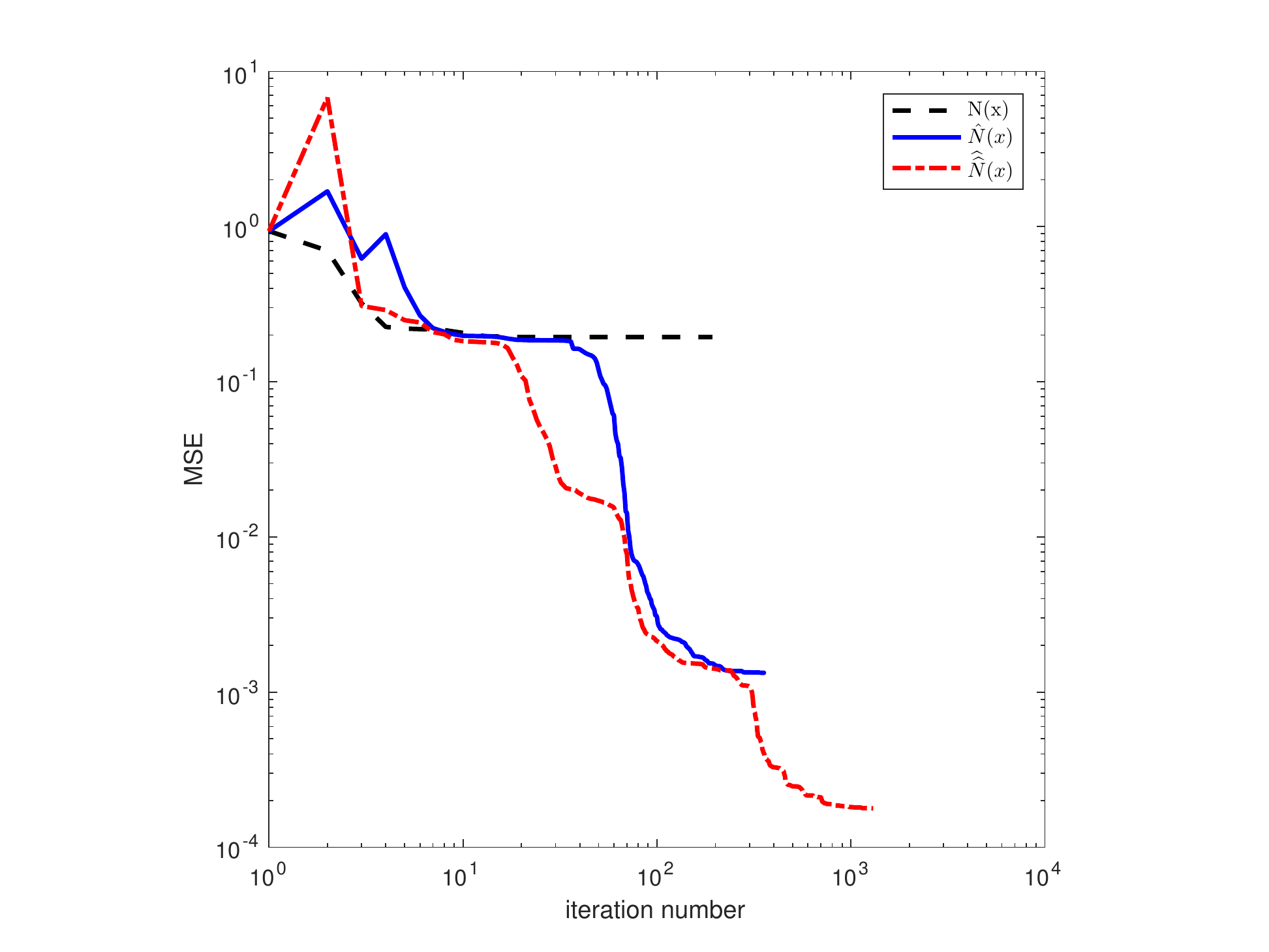} 
    \caption{convergence history} 
  \end{subfigure} 
  \caption{Numerical results for \eqref{testf1} with a second
    sequence of
    random initialization.}
  \label{fig:1L1D_IC2}
\end{figure}

\begin{figure}[ht] 
  \begin{subfigure}[b]{0.5\linewidth}
    \centering
	\includegraphics[width=1\linewidth]{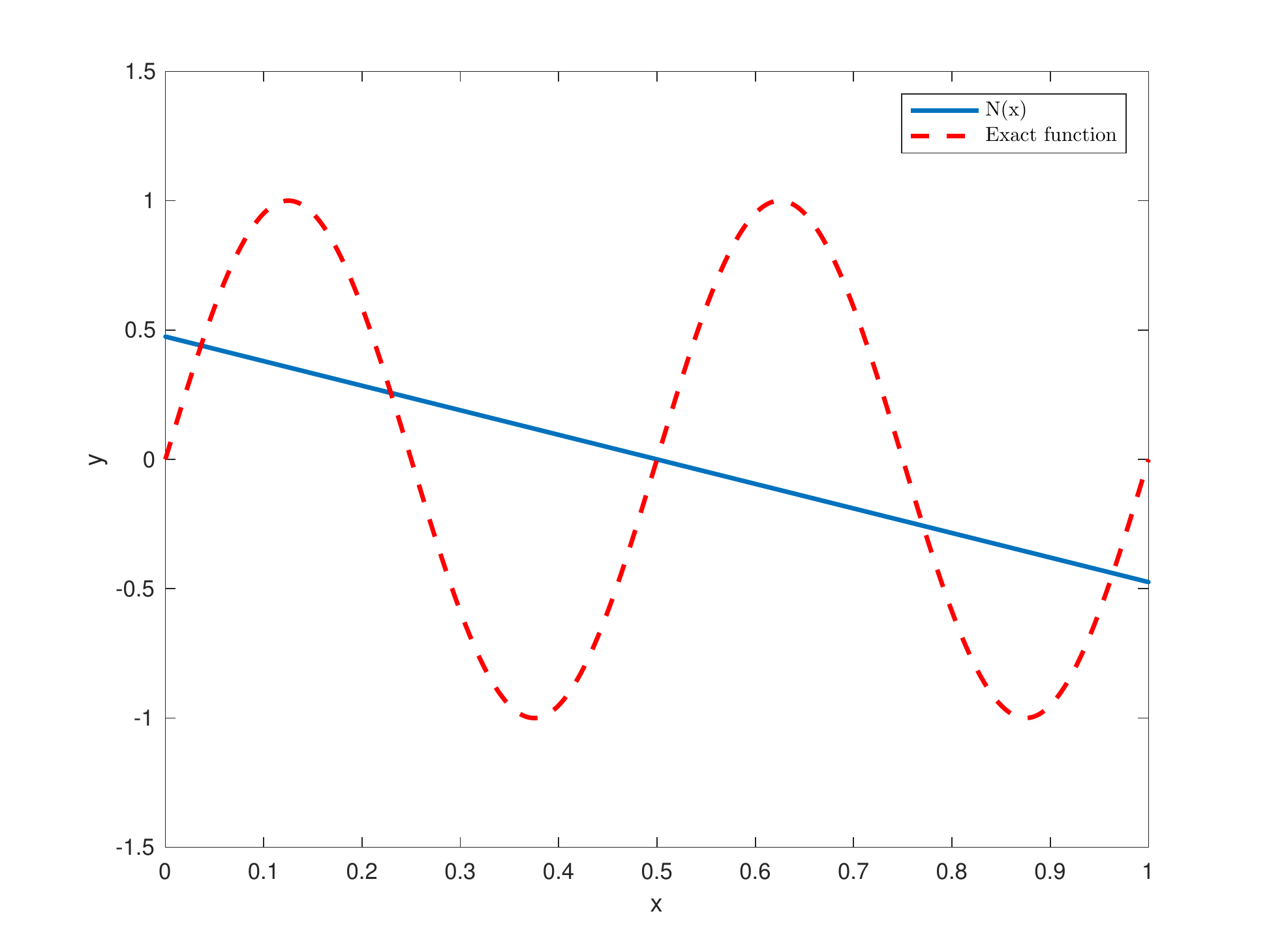} 
	\caption{Unconstrained NN $N(x)$ \eqref{Nx}} 
  \end{subfigure}
  \begin{subfigure}[b]{0.5\linewidth}
    \centering
	\includegraphics[width=1\linewidth]{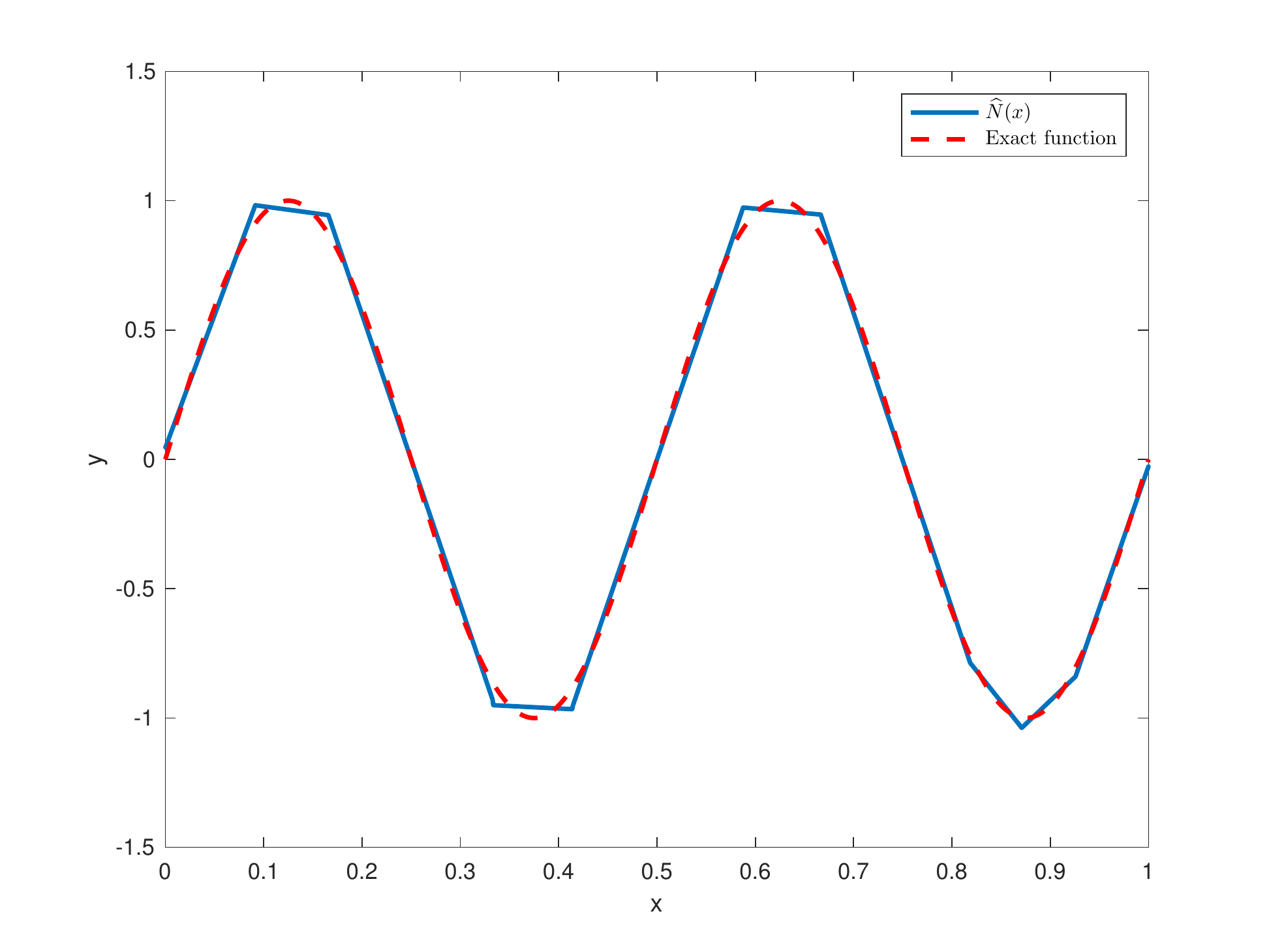} 
	\caption{Constrained NN $\Nh(x)$ \eqref{Nx_1D1}} 
  \end{subfigure} 
  \begin{subfigure}[b]{0.5\linewidth}
    \centering
	\includegraphics[width=1\linewidth]{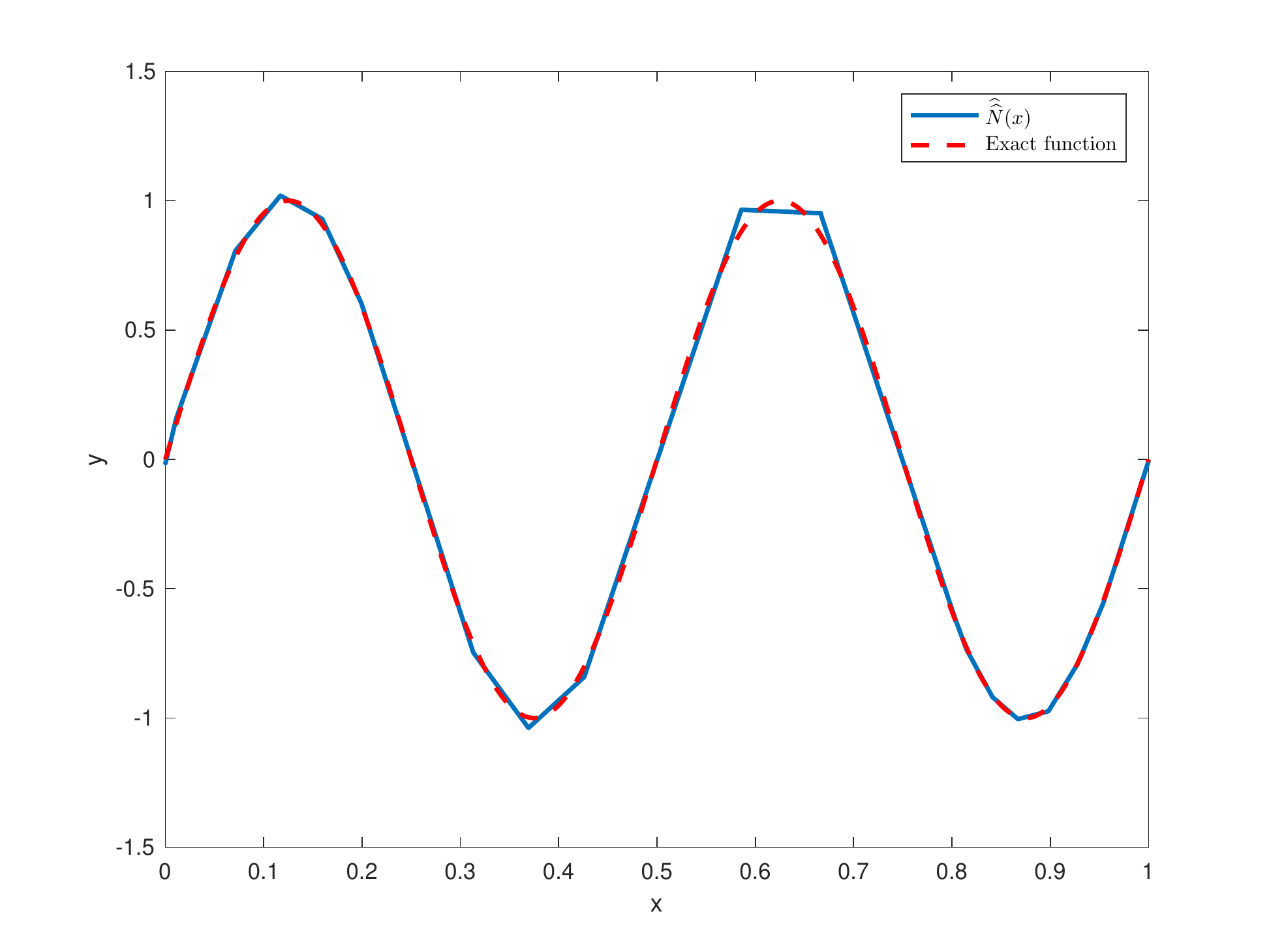} 
	\caption{Constrained NN $\Nhh(x)$ \eqref{Nx_1D2}}
  \end{subfigure}
  \begin{subfigure}[b]{0.5\linewidth}
    \centering
	\includegraphics[width=1\linewidth]{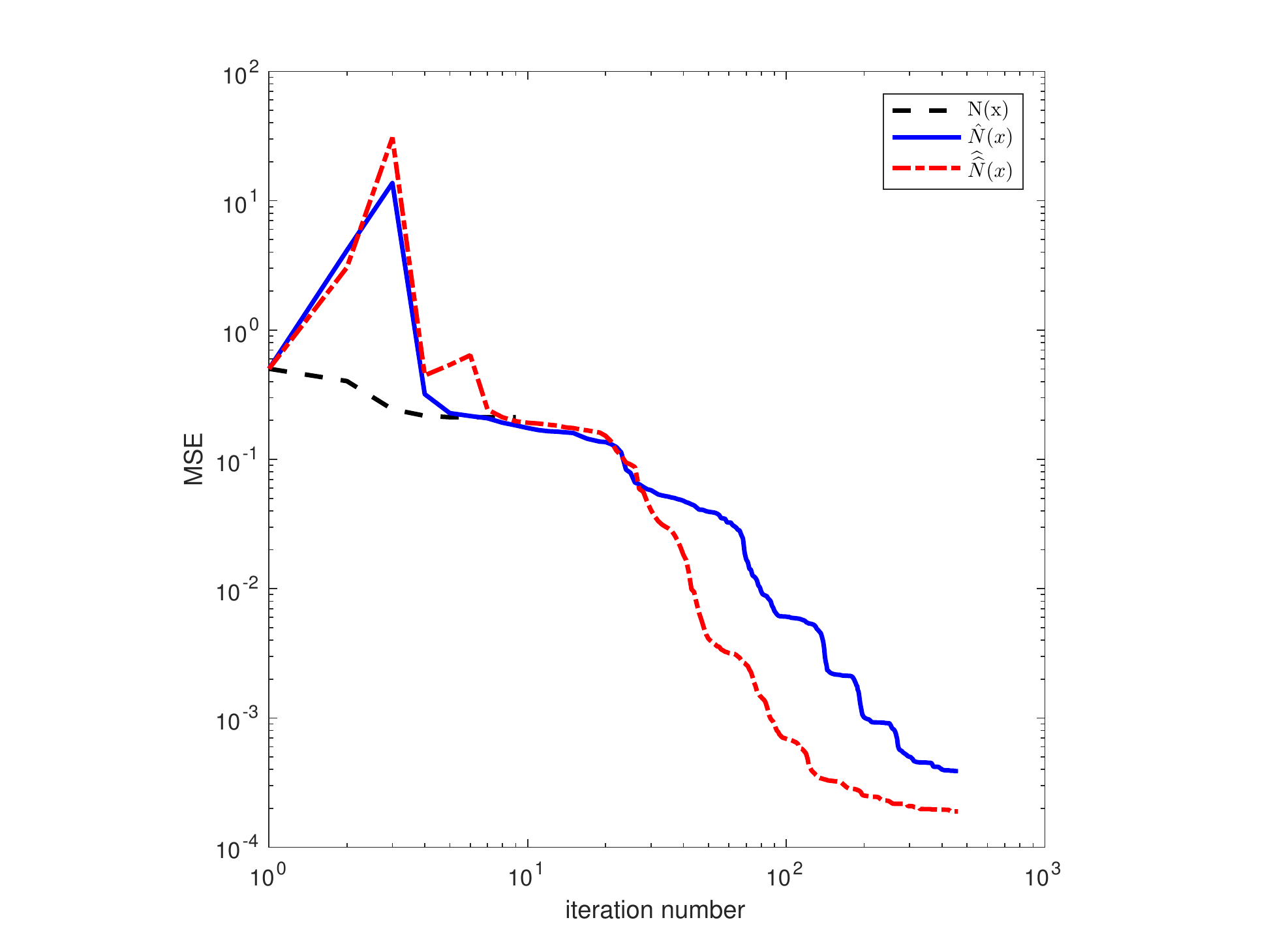} 
    \caption{convergence history} 
  \end{subfigure} 
  \caption{Numerical results for \eqref{testf1} with a third
    sequence of
    random initialization.}
  \label{fig:1L1D_IC3}
\end{figure}

We approximate \eqref{testf1} with NNs with one hidden layer, which consists of
$20$ neurons. The size of the training data set is $200$. Numerical
tests were performed for different choices of random
initializations. It is known that NN training performance depends on
the initialization. In Figures \ref{fig:1L1D_IC1}, \ref{fig:1L1D_IC2} and
\ref{fig:1L1D_IC3}, we show the numerical results for three sets of
different random initializations. In each set, the unconstrained NN
\eqref{Nx}, the constrained NN \eqref{Nx_1D1} and the specialized
constrained NN with \eqref{Nx_1D2} use the same random sequence for
initialization.
We observe that the standard NN formulation without constraints
\eqref{Nx} produces training results critically dependent on the
initialization. This is widely acknowledged in the literature. On the other hand, our new
constrained NN formulations are more robust and produce better
results that are less sensitive to the initialization. The tighter constraint \eqref{Nx_1D2} performs better than
the general constraint \eqref{Nx_1D1}, which is not
surprising. However, the tighter constraint is a special case for this
particular problem and not available in the general case.

\subsubsection{Two-dimensional tests}

\begin{figure}[ht] 
  \begin{subfigure}[b]{0.5\linewidth}
    \centering
	\includegraphics[width=1\linewidth]{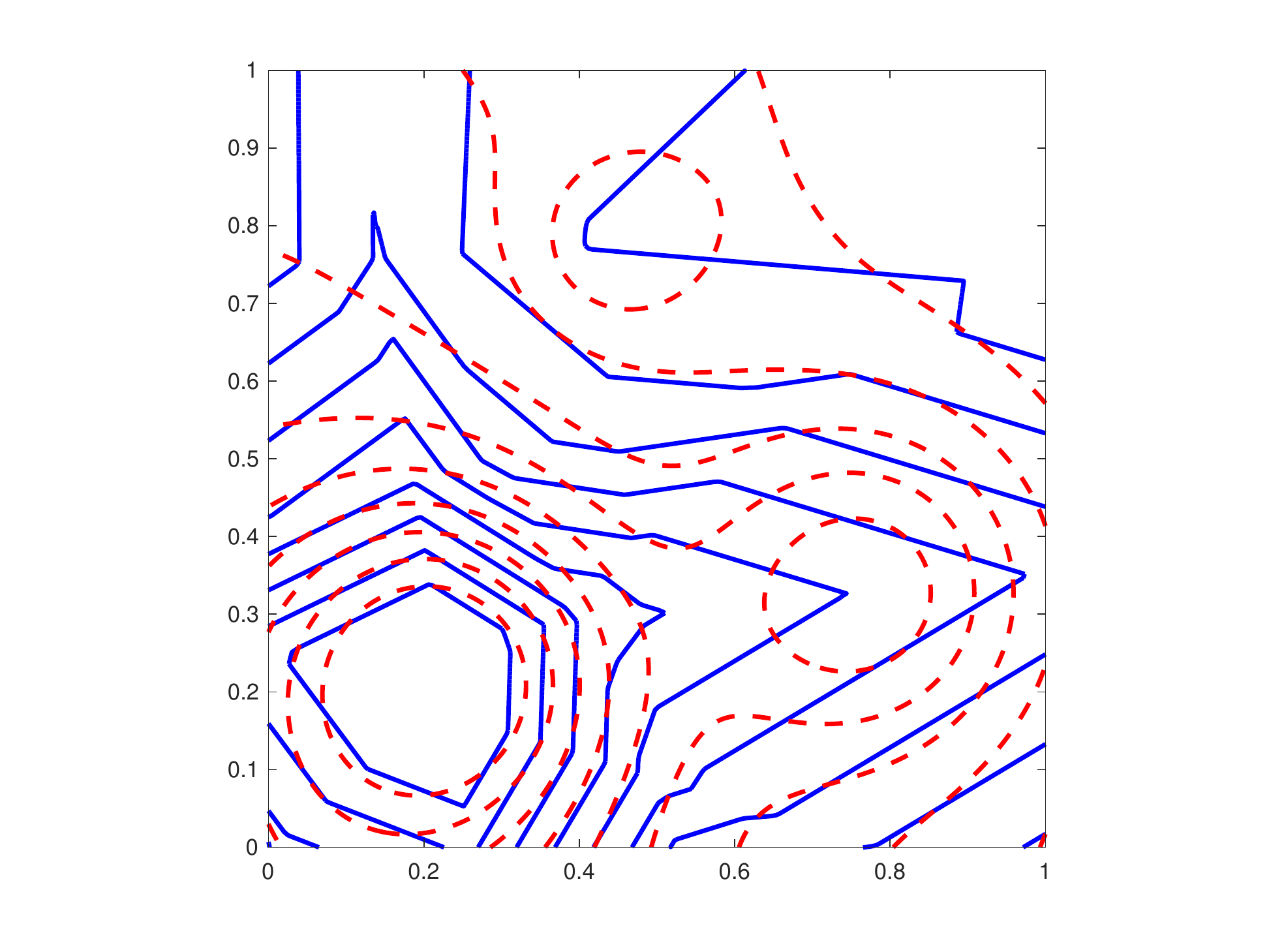} 
	\caption{Unconstrained \eqref{Nx}: contours} 
  \end{subfigure}
	\begin{subfigure}[b]{0.5\linewidth}
    \centering
	\includegraphics[width=1\linewidth]{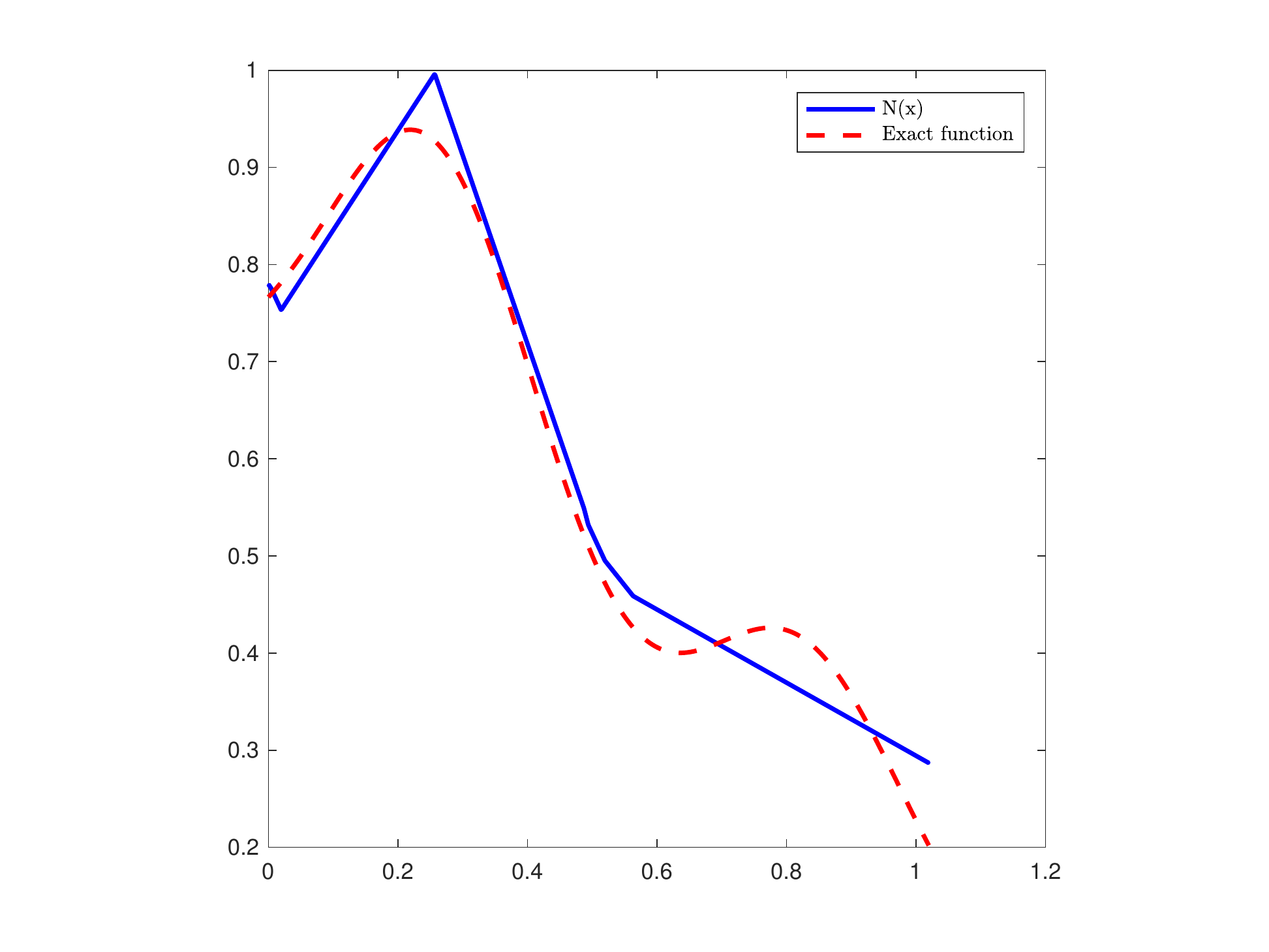} 
	\caption{Unconstrained \eqref{Nx}: $y=0.2x$ cut} 
  \end{subfigure} 
  \begin{subfigure}[b]{0.5\linewidth}
    \centering
	\includegraphics[width=1\linewidth]{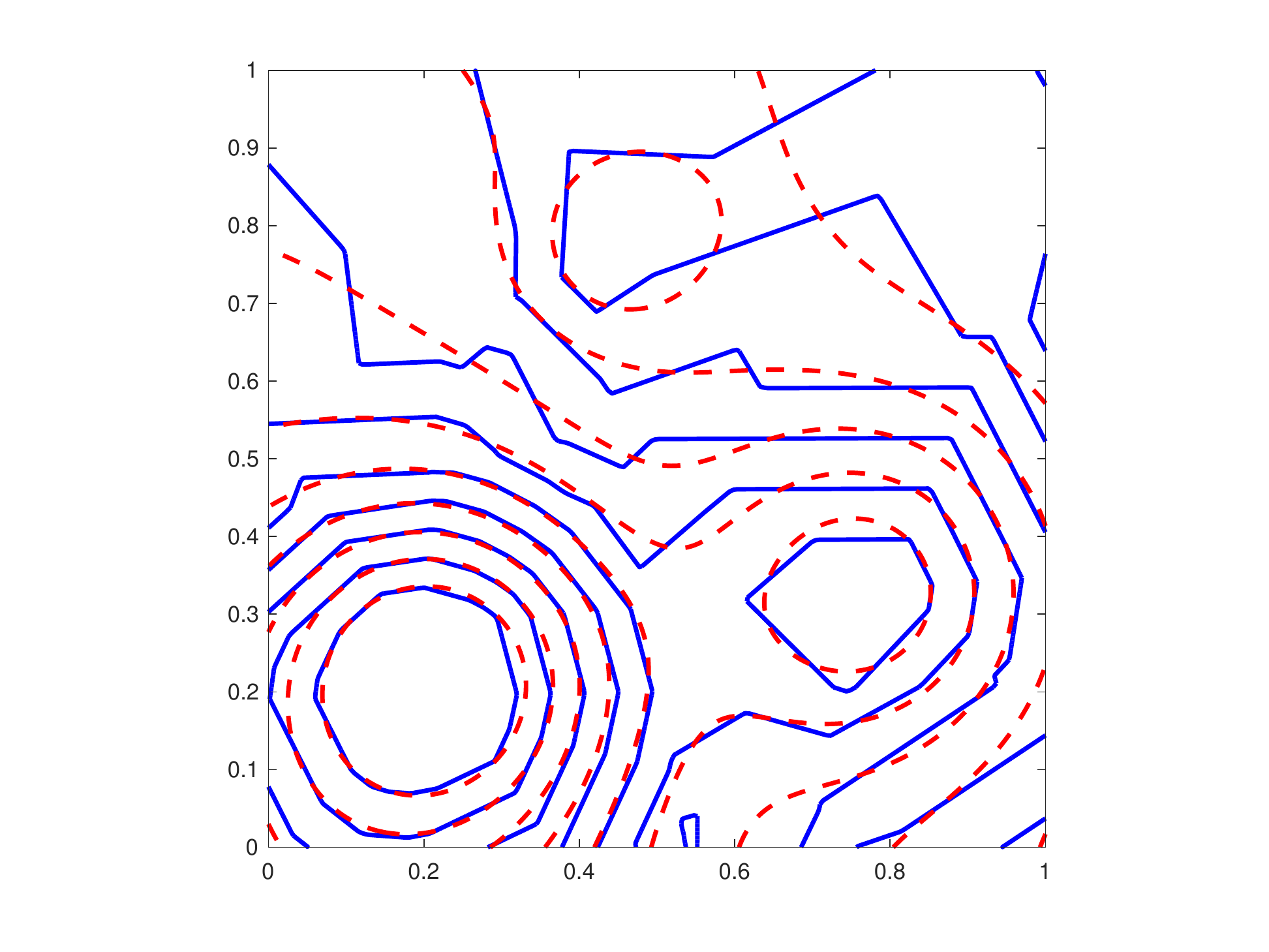} 
	\caption{Constrained \eqref{Nnew2}: contours} 
  \end{subfigure} 
  \begin{subfigure}[b]{0.5\linewidth}
    \centering
	\includegraphics[width=1\linewidth]{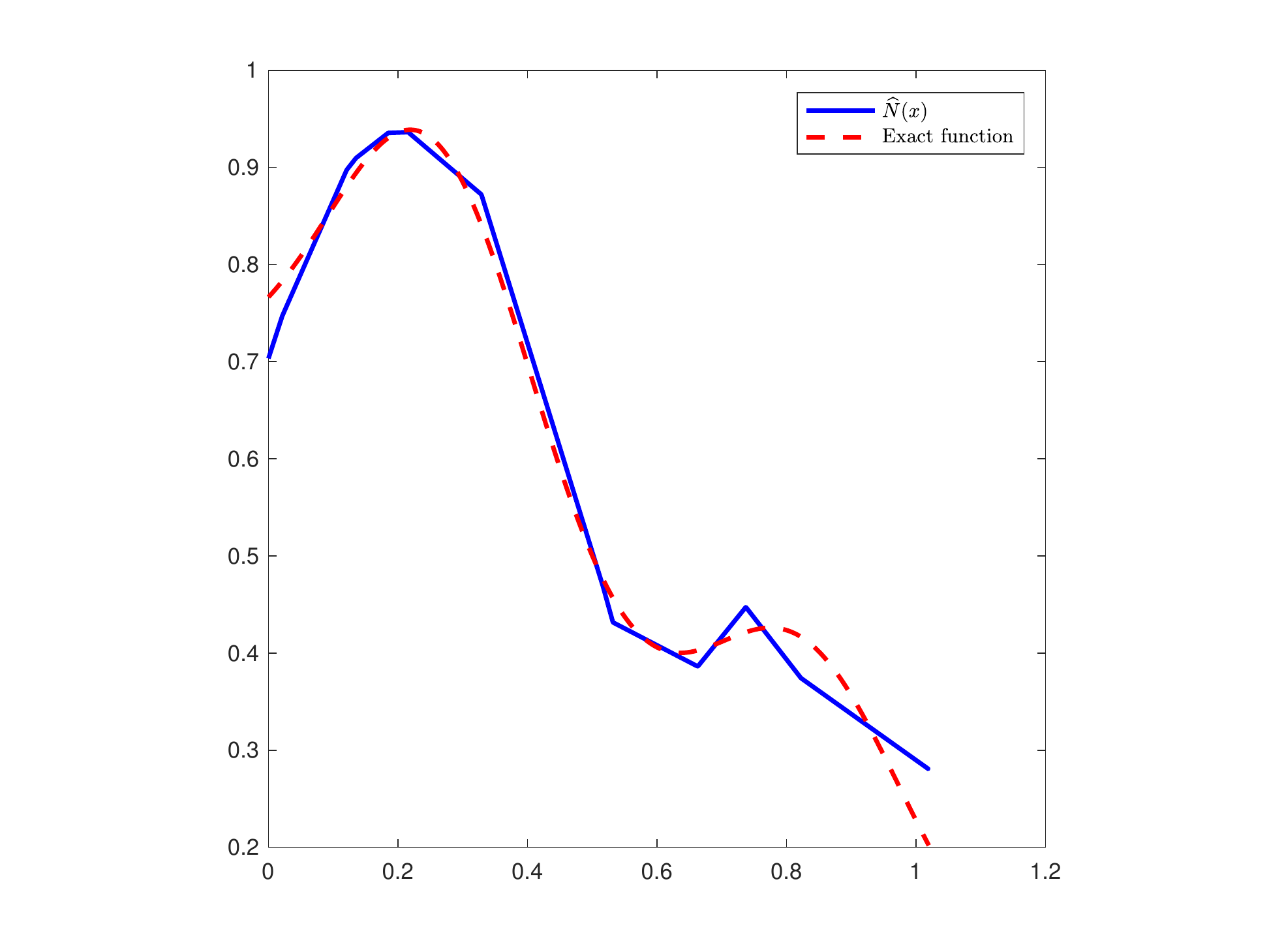} 
	\caption{Constrained \eqref{Nnew2}: $y=0.2x$ cut}
  \end{subfigure}
  \caption{Numerical results for \eqref{testf2}. Top row: unconstrained
    formulation $N(x)$ \eqref{Nx}; Bottom row: constrained
    formulation $\Nh(x)$ \eqref{Nnew2}.}
\label{fig:1L2D}
\end{figure}

We next consider two-dimensional functions. In
particular, we show result for the Franke's function \cite{franke1982}
\begin{equation}\label{testf2}
	\begin{split}
		f(x,y)&=\frac{3}{4}\exp\left(-\frac{(9x-2)^2}{4}-\frac{(9y-2)^2}{4}\right)+\frac{3}{4} \exp\left(-\frac{(9x+1)^2}{49}-\frac{9y+1}{10}\right)\\
	&+\frac{1}{2}\exp\left(-\frac{(9x-7)^2}{4}-\frac{(9y-3)^2}{4}\right)-\frac{1}{5}\exp\left(-(9x-4)^2-(9y-7)^2\right)
	\end{split}
\end{equation}
with $(x, y)\in [0,1]^2$.
Again, we compare training results for both the standard NN without
constraints \eqref{Nx} and our new constrained NN formulation
\eqref{Nnew}, using the same random sequence for initialization.
The NNs have one
hidden layer with $40$ neurons. The size of the training set is $500$ and that
of the validation set is $1,000$. 
The numerical results are shown in  Figure \ref{fig:1L2D}. On the left
column, the contour lines of the training results are shown, as well
as those of the exact function. Here all contour lines are at the same
values, from 0 to 1 with an increment of $0.1$.
We observe that the constrained NN formulation produces visually
better result than the standard unconstrained formulation. On the
right column, we plot the function value along $y=0.2 x$. Again, the
improvement of the constrained NN is visible.

\subsection{Multiple Hidden Layers}

We now consider feedforward NNs with multiple hidden layers. We present results
for both
the standard NN without constraints \eqref{MultNx} and the
constrained ReLU NNs with the constraints \eqref{weight_general} and
\eqref{b_general}. We use the standard notation $\{J_1,\dots, J_M\}$
to denote the network structure, where $J_m$ is the number of neurons
in each layer. The hidden layers are $J_2,\dots, J_{M-1}$.
Again, we focus on 1D and 2D functions for ease of
visualization purpose, i.e., $J_1=1, 2$.

\subsubsection{One dimensional tests}

\begin{figure}[ht] 
  
  \begin{subfigure}[b]{0.5\linewidth}
    \centering
	\includegraphics[width=1\linewidth]{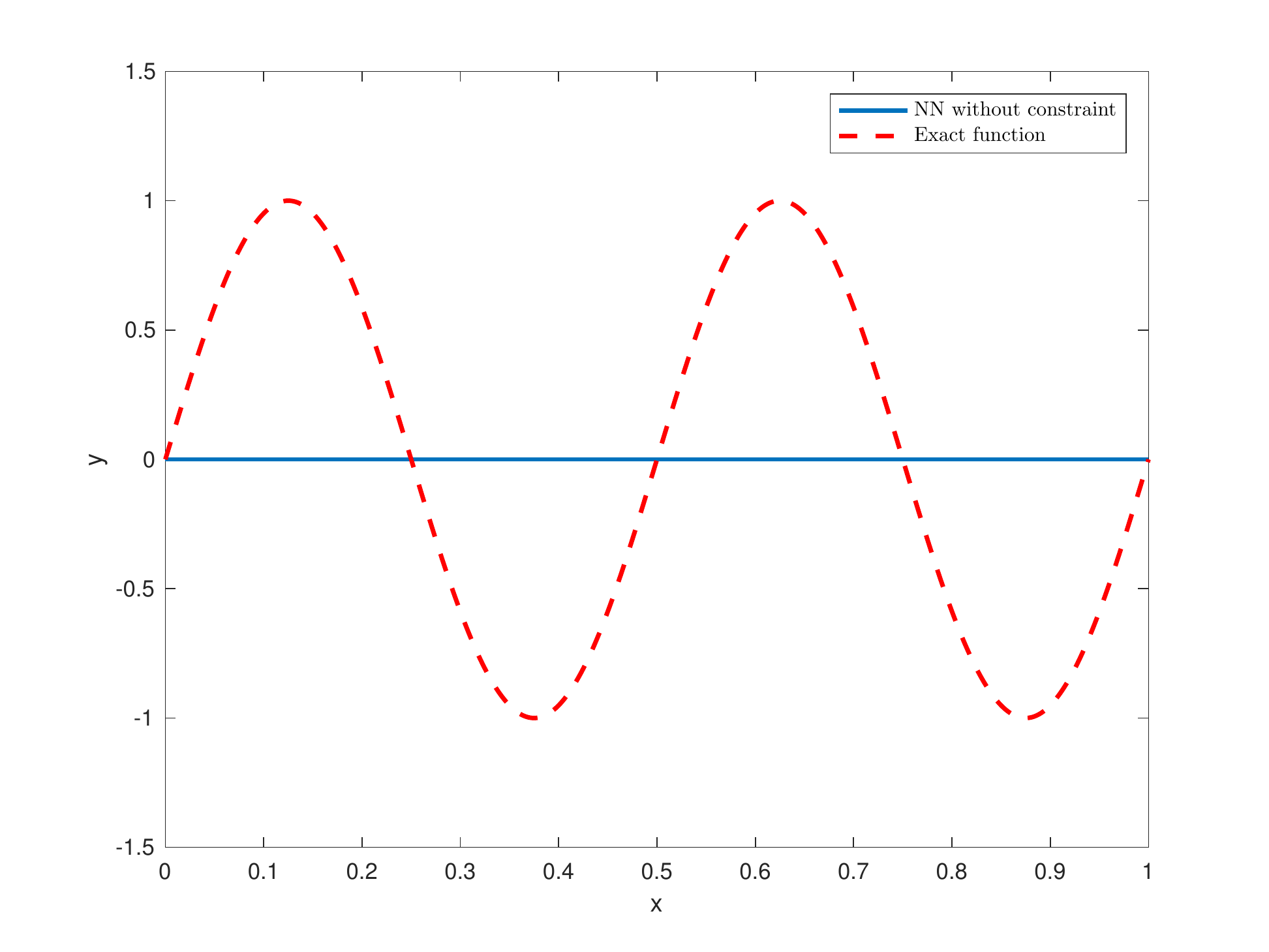} 
	\caption{Unconstrained NN}
  \end{subfigure}
  \begin{subfigure}[b]{0.5\linewidth}
    \centering
	\includegraphics[width=1\linewidth]{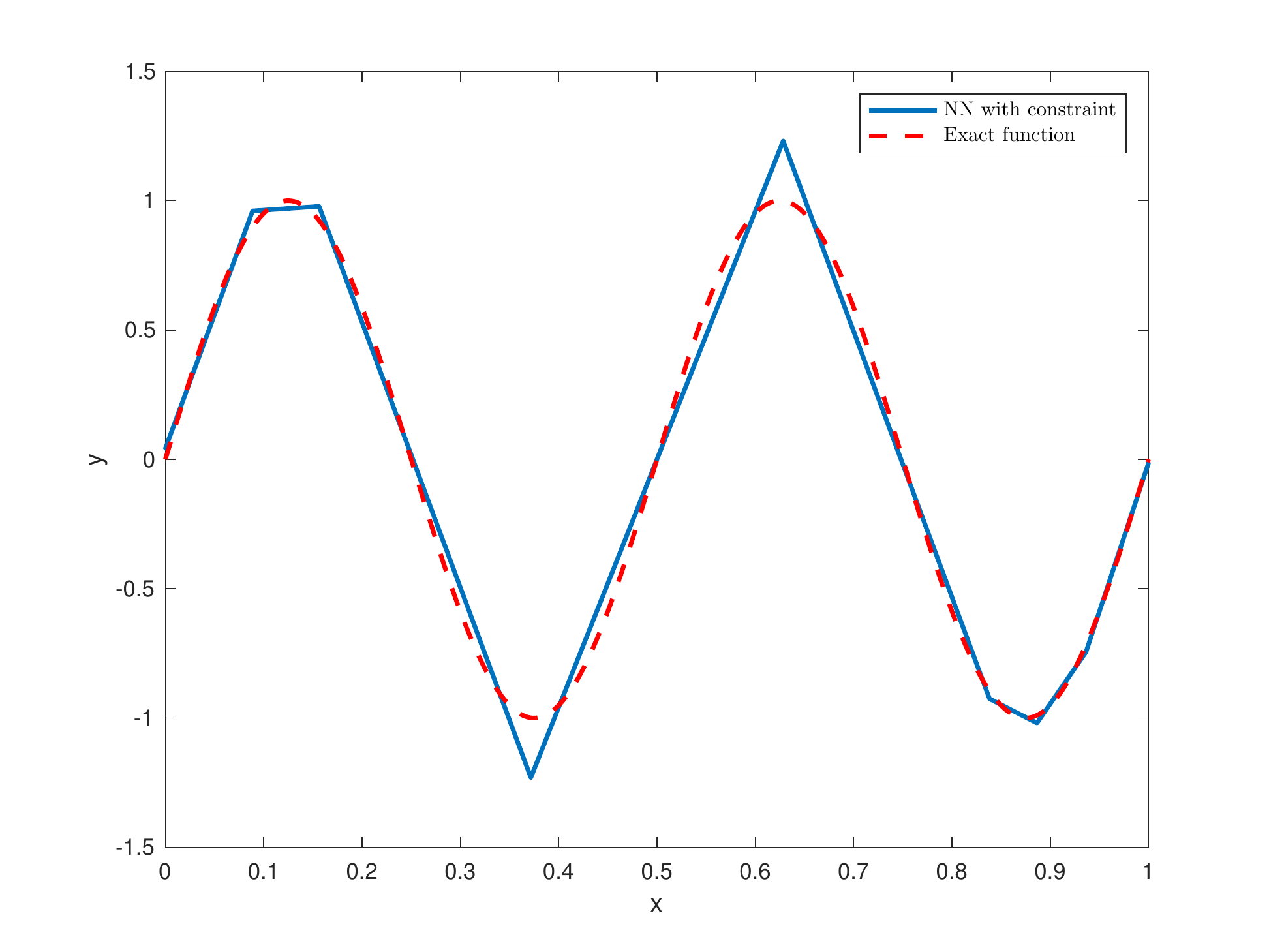} 
	\caption{Constrained NN with \eqref{weight_general} and \eqref{b_general}} 
  \end{subfigure} 
  \begin{subfigure}[b]{0.5\linewidth}
    \centering
	\includegraphics[width=1\linewidth]{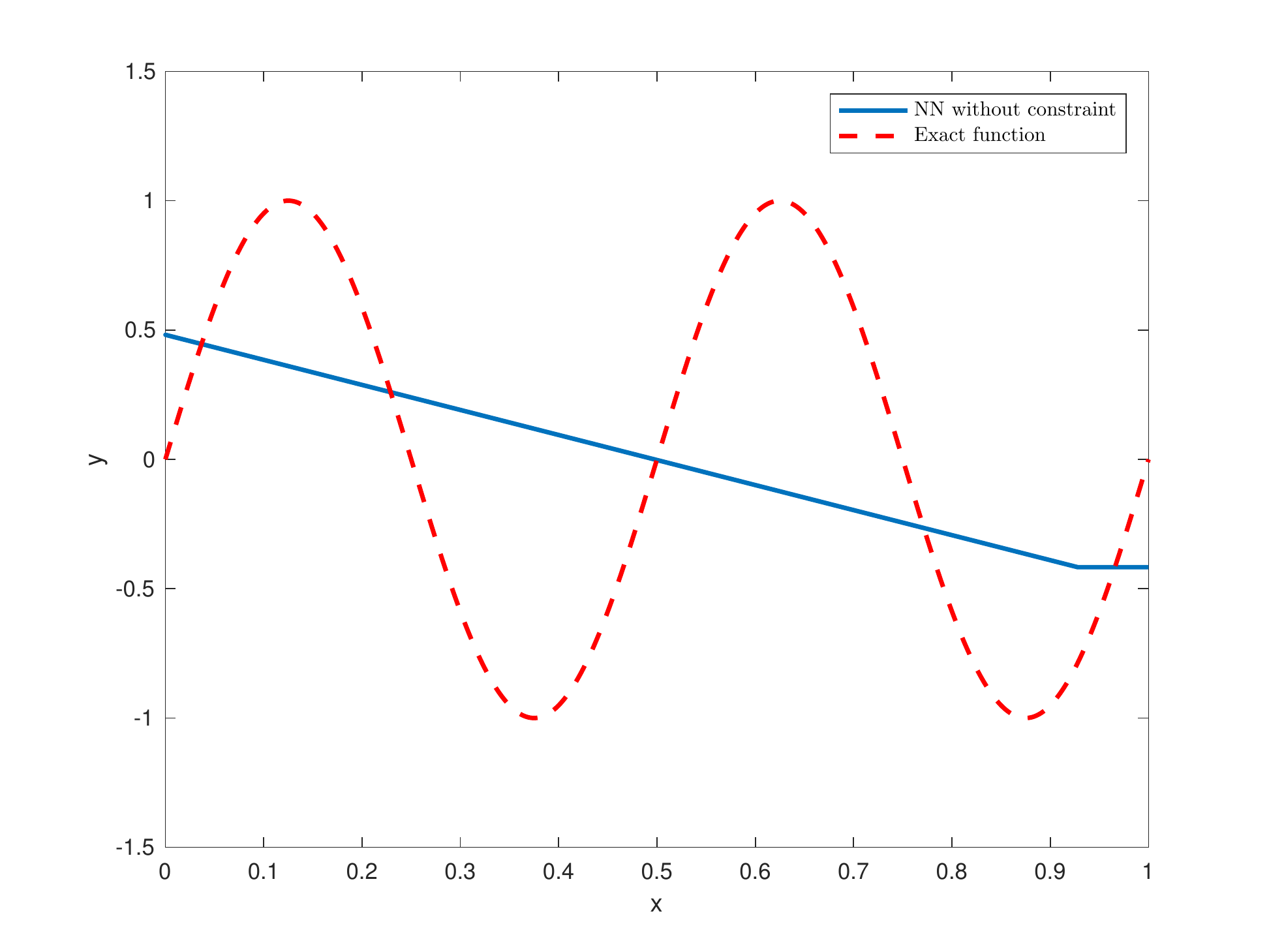} 
    \caption{Unconstrained NN} 
  \end{subfigure} 
  \begin{subfigure}[b]{0.5\linewidth}
    \centering
	\includegraphics[width=1\linewidth]{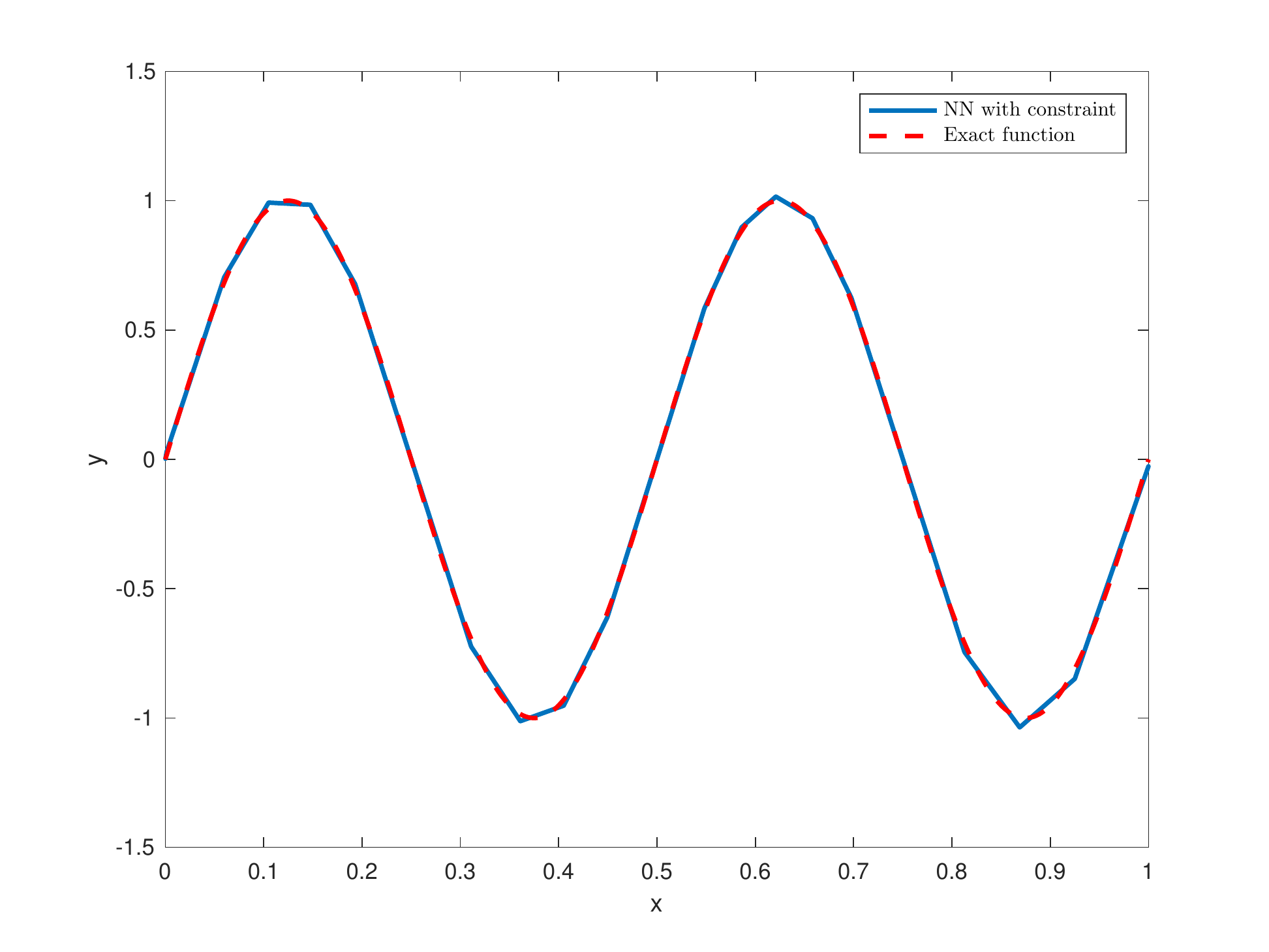} 
	\caption{Constrained NN with \eqref{weight_general} and \eqref{b_general}}
  \end{subfigure} 
  \begin{subfigure}[b]{0.5\linewidth}
    \centering
	\includegraphics[width=1\linewidth]{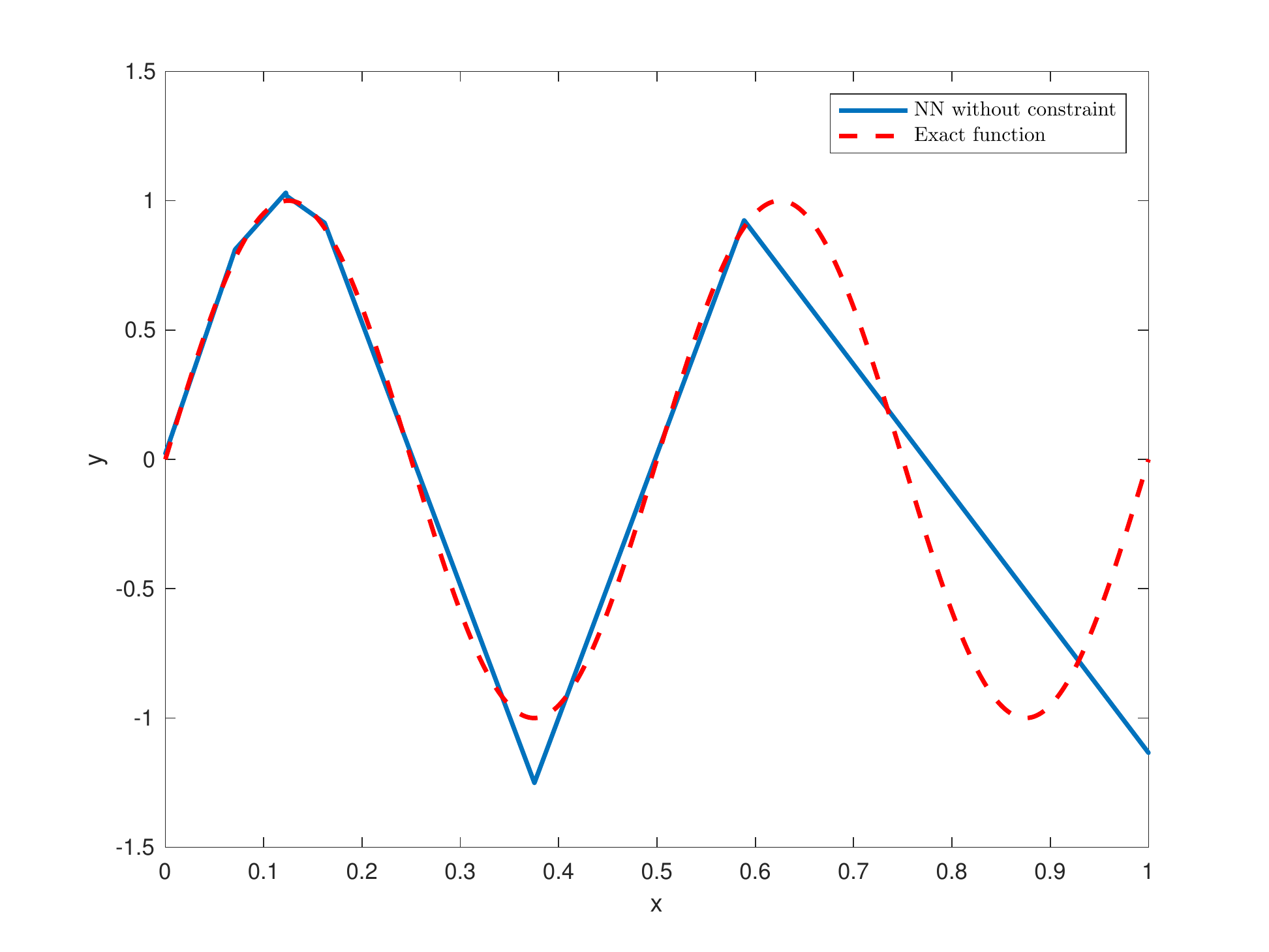} 
    \caption{Unconstrained NN} 
  \end{subfigure} 
  \begin{subfigure}[b]{0.5\linewidth}
    \centering
	\includegraphics[width=1\linewidth]{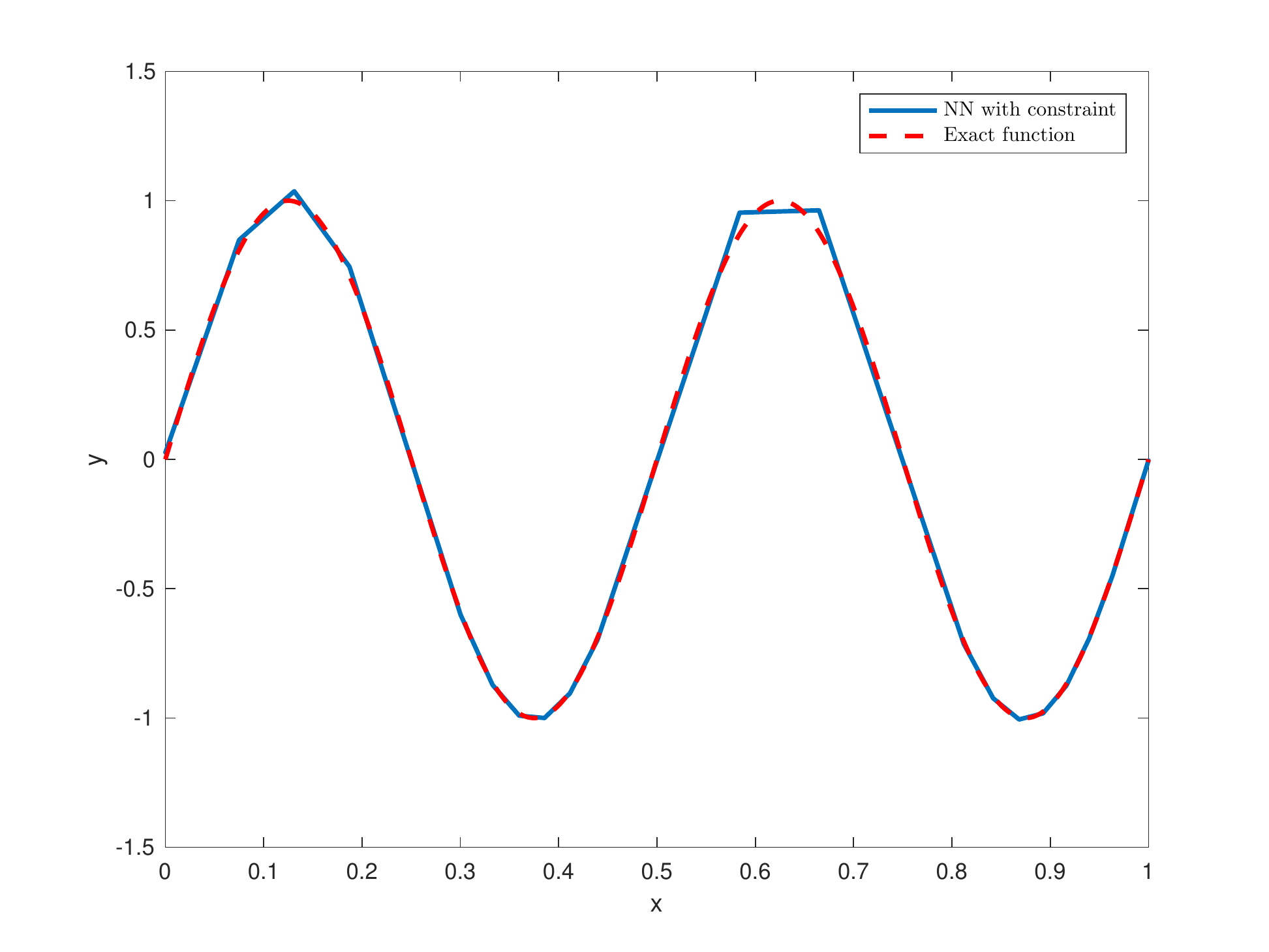} 
	\caption{Constrained NN with \eqref{weight_general} and \eqref{b_general}} 
  \end{subfigure} 
  \caption{Numerical results for 1D function \eqref{testf1} with
    feedforward NNs of $\{1, 20, 10, 1\}$. From top to bottom:
    training results using three different random sequences for initialization.
 Left column: results by unconstrained NN formulation \eqref{MultNx};
 Right column: results by NN formulation with constraints \eqref{weight_general} and \eqref{b_general}.}
  \label{fig:2L1D}
\end{figure}

\begin{figure}[ht] 
    \begin{subfigure}[b]{0.5\linewidth}
    \centering
	\includegraphics[width=1\linewidth]{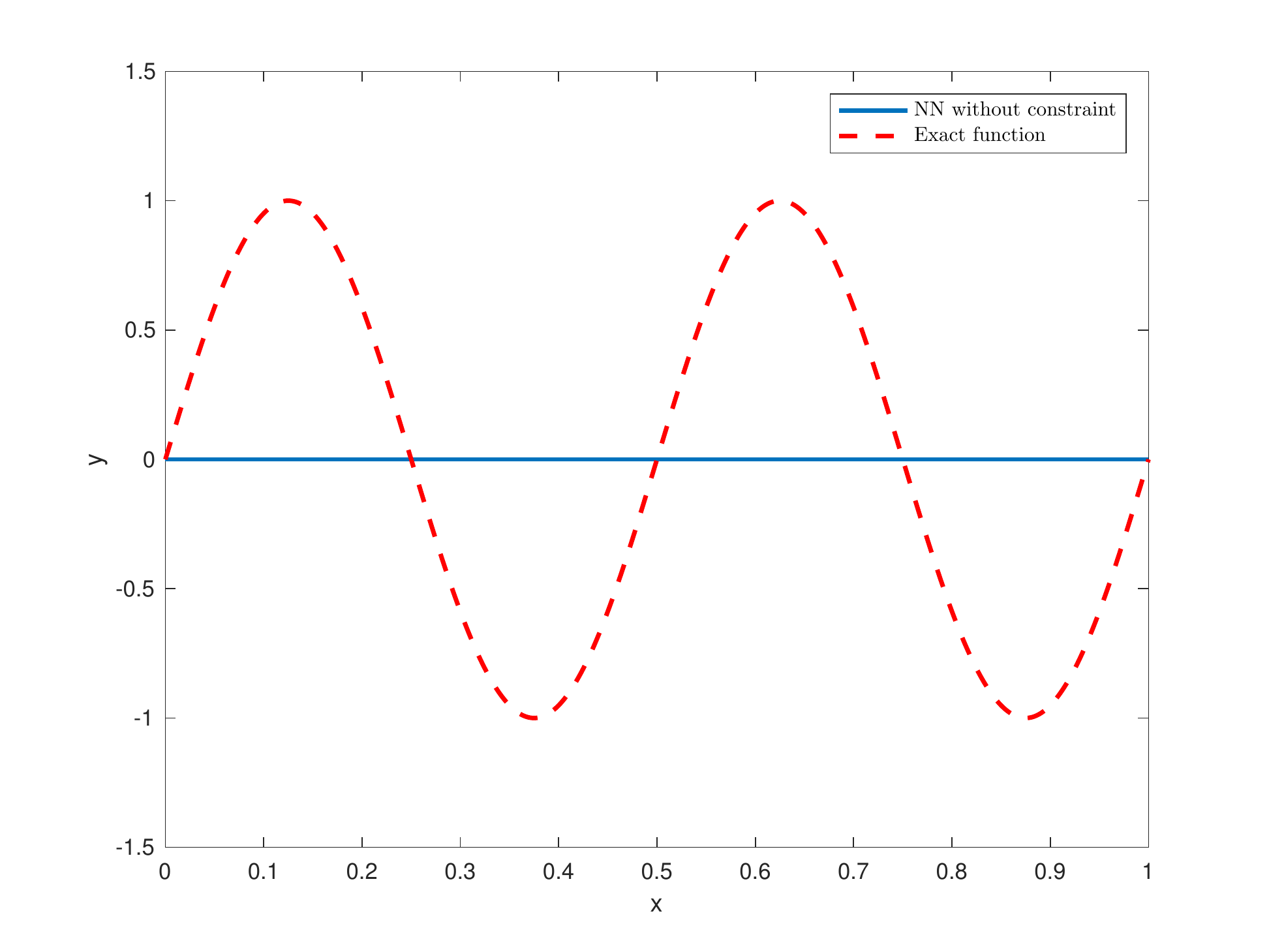} 
	\caption{Unconstrained NN} 
  \end{subfigure}
  \begin{subfigure}[b]{0.5\linewidth}
    \centering
	\includegraphics[width=1\linewidth]{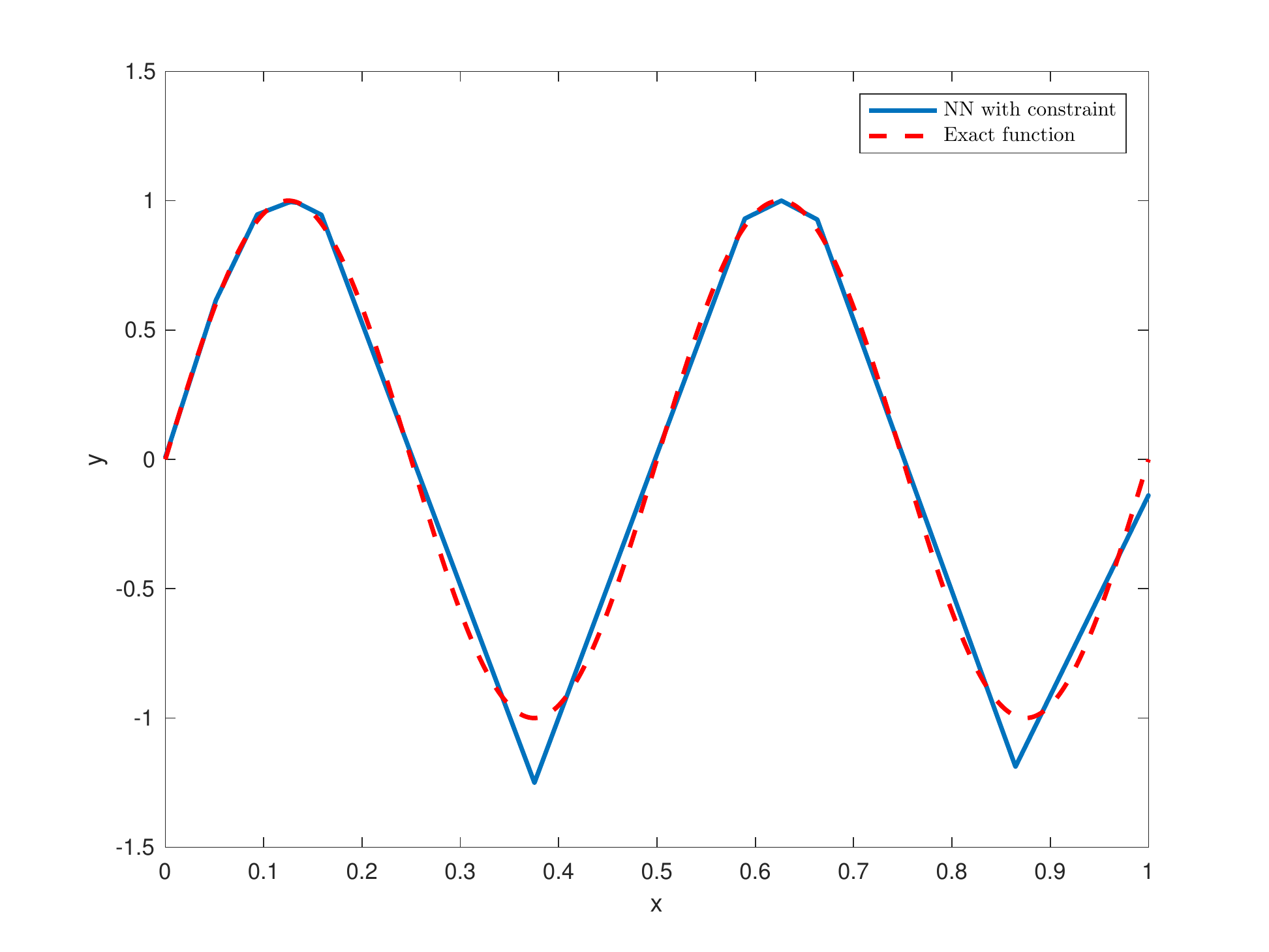} 
	\caption{Constrained NN with \eqref{weight_general} and \eqref{b_general}} 
  \end{subfigure} 
  \begin{subfigure}[b]{0.5\linewidth}
    \centering
	\includegraphics[width=1\linewidth]{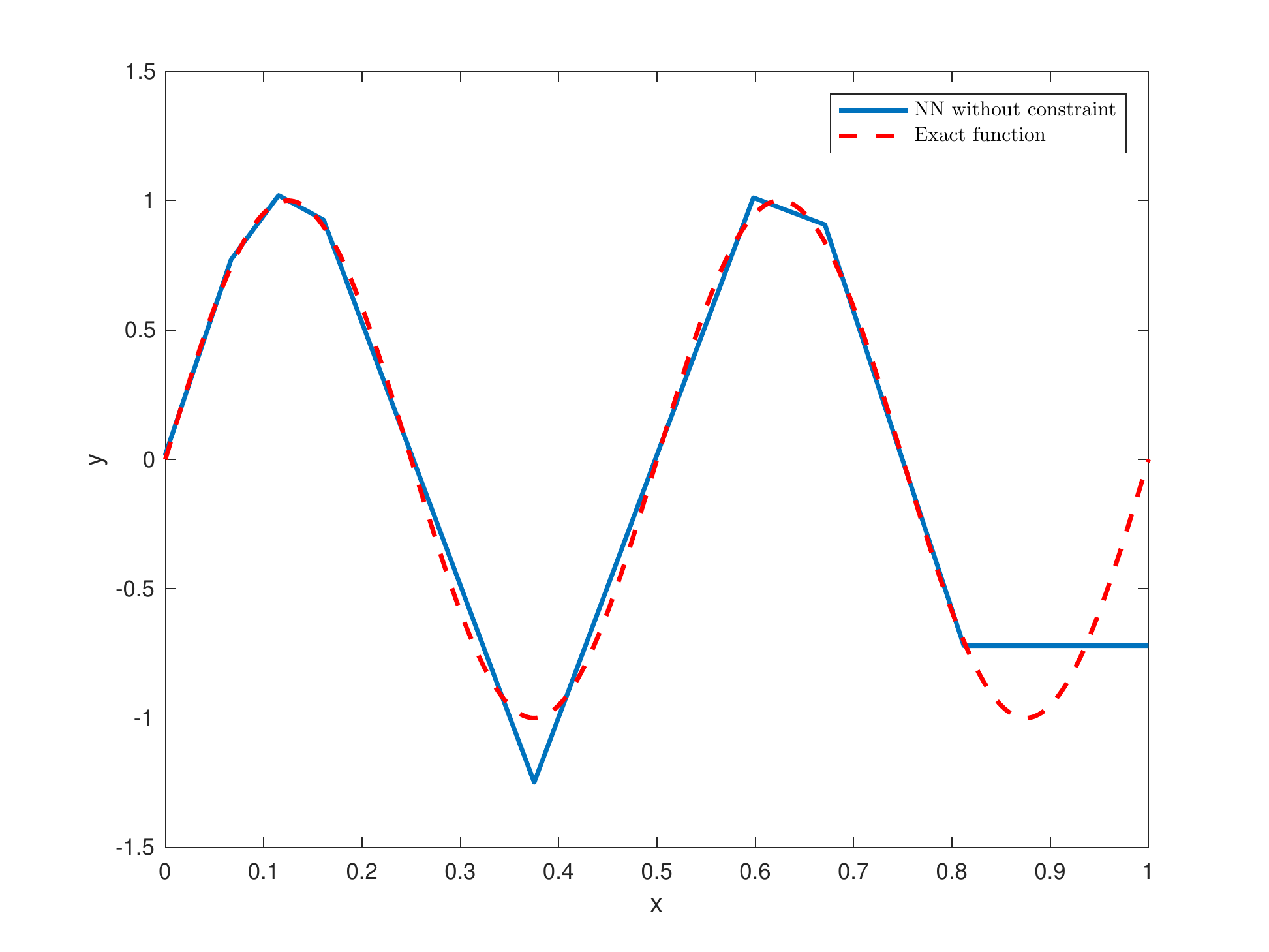} 
    \caption{Unconstrained NN} 
  \end{subfigure} 
  \begin{subfigure}[b]{0.5\linewidth}
    \centering
	\includegraphics[width=1\linewidth]{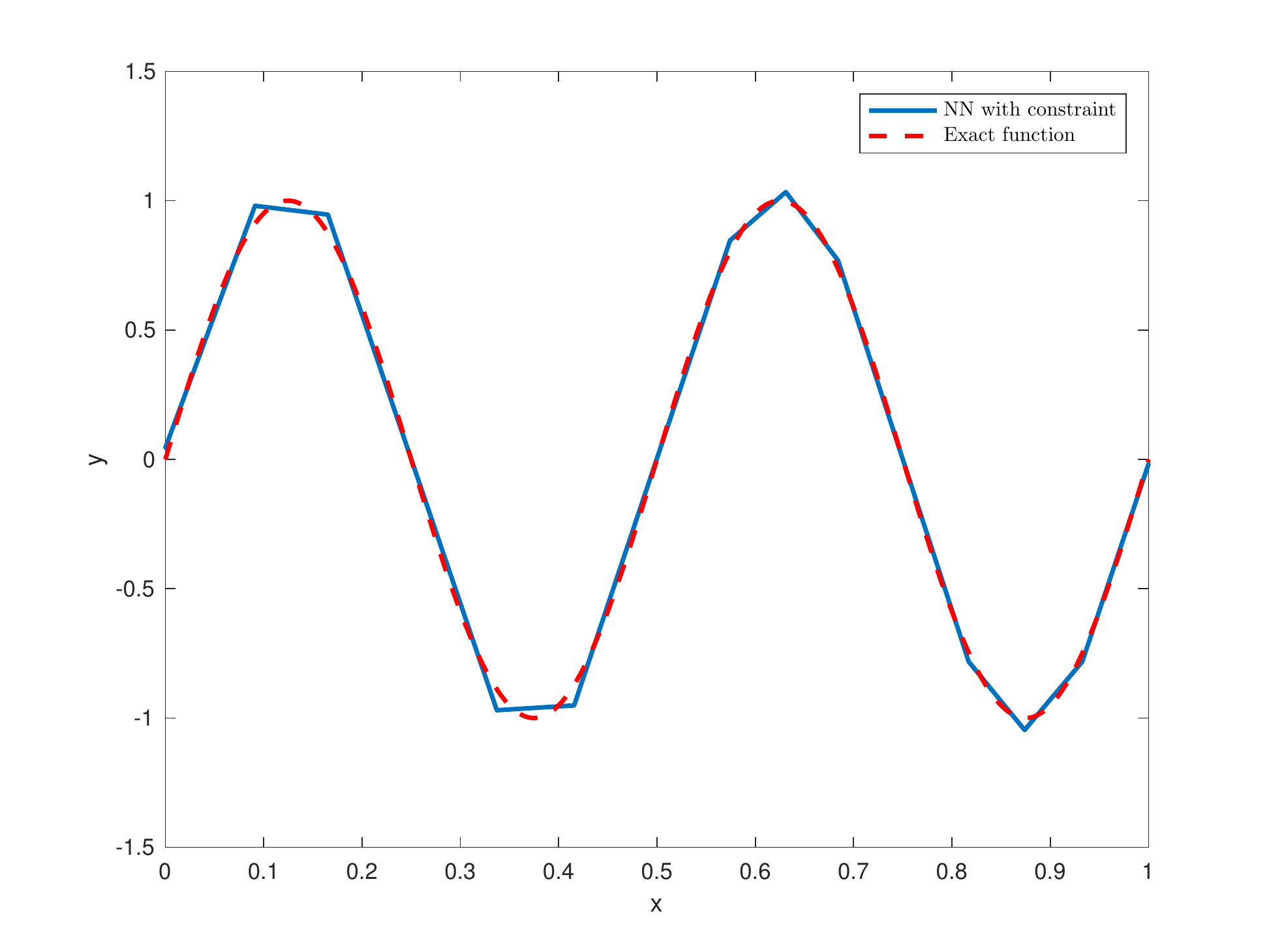} 
	\caption{Constrained NN with \eqref{weight_general} and \eqref{b_general}}
  \end{subfigure} 
  \begin{subfigure}[b]{0.5\linewidth}
    \centering
	\includegraphics[width=1\linewidth]{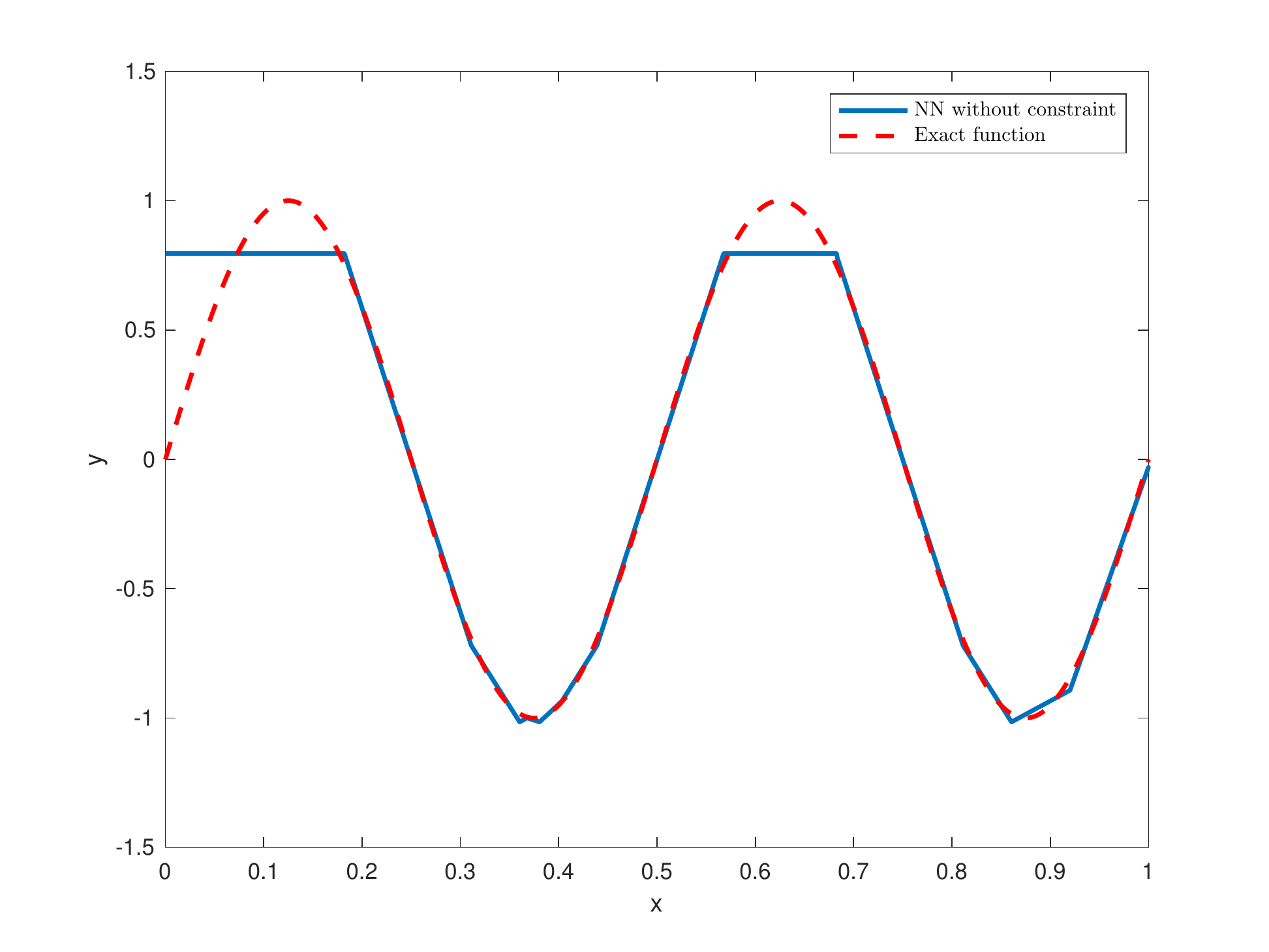} 
    \caption{Unconstrained NN} 
  \end{subfigure} 
  \begin{subfigure}[b]{0.5\linewidth}
    \centering
	\includegraphics[width=1\linewidth]{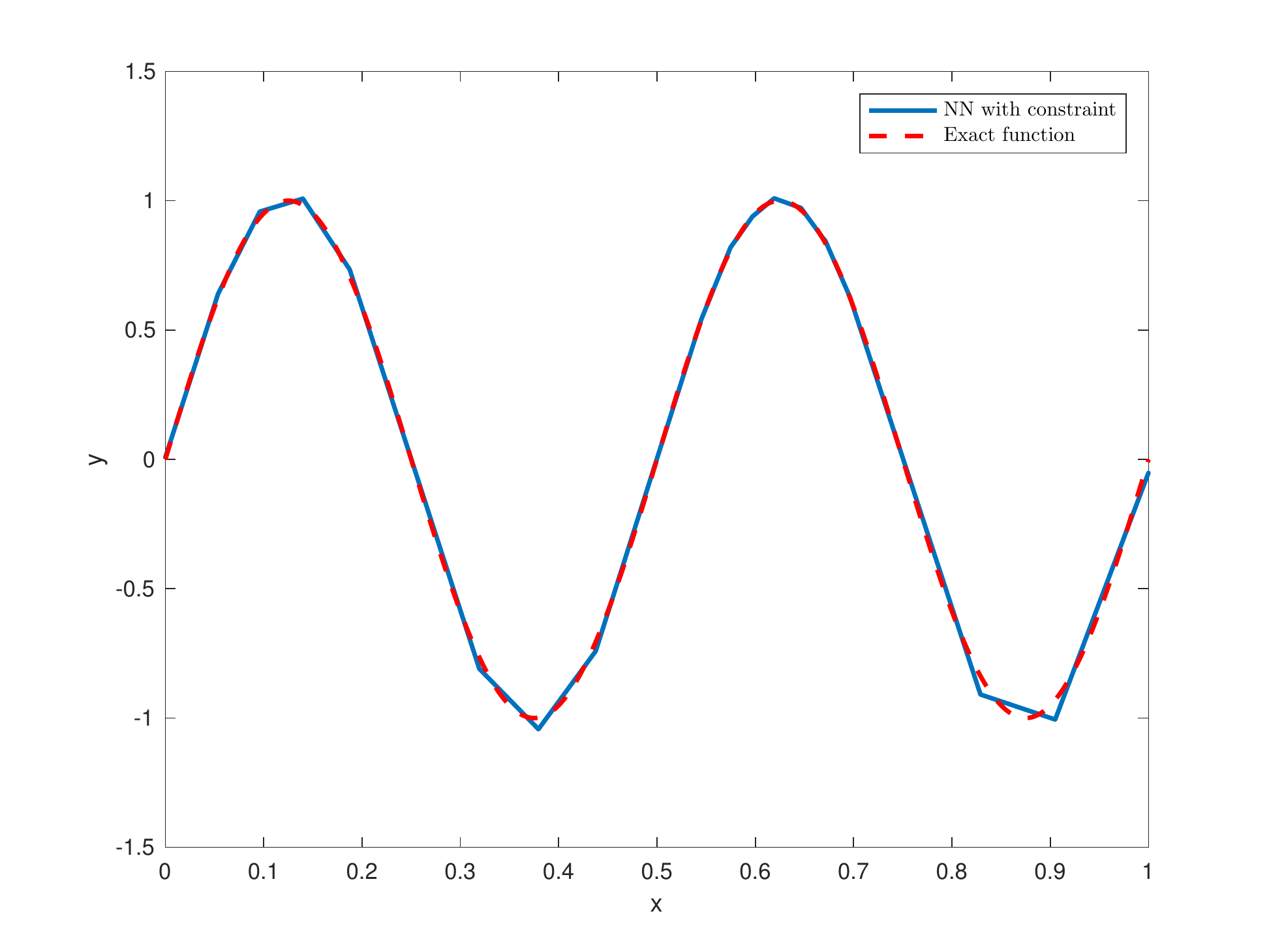} 
	\caption{Constrained NN with \eqref{weight_general} and \eqref{b_general}} 
  \end{subfigure} 
  \caption{Numerical results for 1D function \eqref{testf1} with
    feedforward NNs of $\{1, 5, 5, 5, 5, 1\}$. From top to bottom:
    training results using three different random sequences for initialization.
 Left column: results by unconstrained NN formulation \eqref{MultNx};
 Right column: results by NN formulation with constraints
 \eqref{weight_general} and \eqref{b_general}.}
  \label{fig:4L1D}
\end{figure}

We first consider the one-dimensional function \eqref{testf1}.
%
In Figure \ref{fig:2L1D}, we show the numerical results by NNs of
$\{1, 20, 10, 1\}$, using three different sequences of random
initializations, with and without constraints. We observe that the
standard NN formulation without constraints \eqref{MultNx} produces widely
different results. This is because of the potentially large number of
local minima in the cost function and is not entirely surprising. On
the other hand, using the exactly the same initialization, the
NN formulation with constraints \eqref{weight_general} and \eqref{b_general} produces notably better
results, and more importantly, is much less sensitive to the
initialization. 
In Figure \ref{fig:4L1D}, we show the results for NNs with
$\{1,5,5,5,5,1\}$ structure. We observe similar performance -- the
constrained NN produces better results and is less sensitive to
initialization.

\subsubsection{Two dimensional tests}

\begin{figure}[ht] 
  \begin{subfigure}[b]{0.5\linewidth}
    \centering
	\includegraphics[width=1\linewidth]{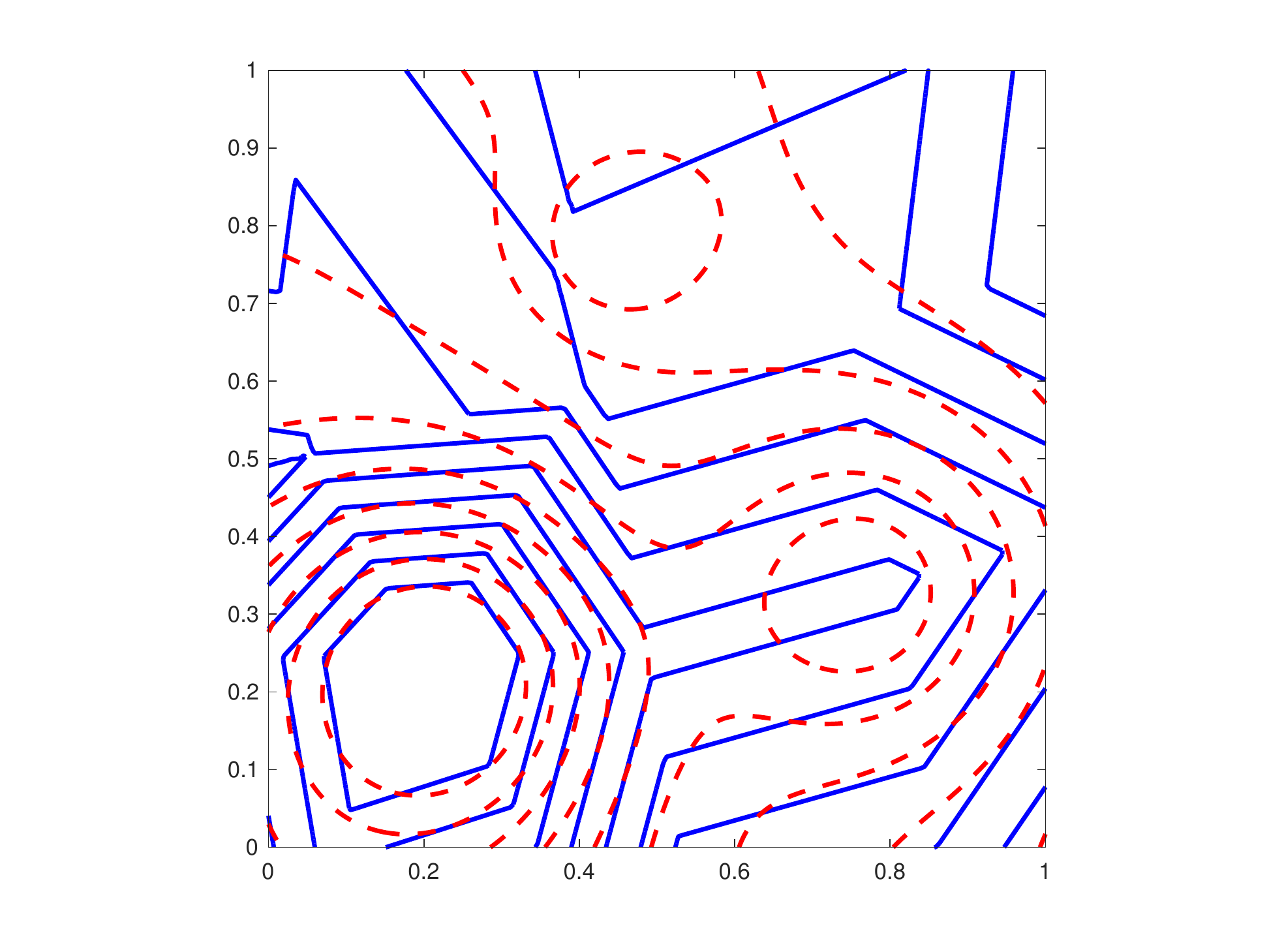} 
	\caption{Unconstrained: contours} 
  \end{subfigure}
	\begin{subfigure}[b]{0.5\linewidth}
    \centering
	\includegraphics[width=1\linewidth]{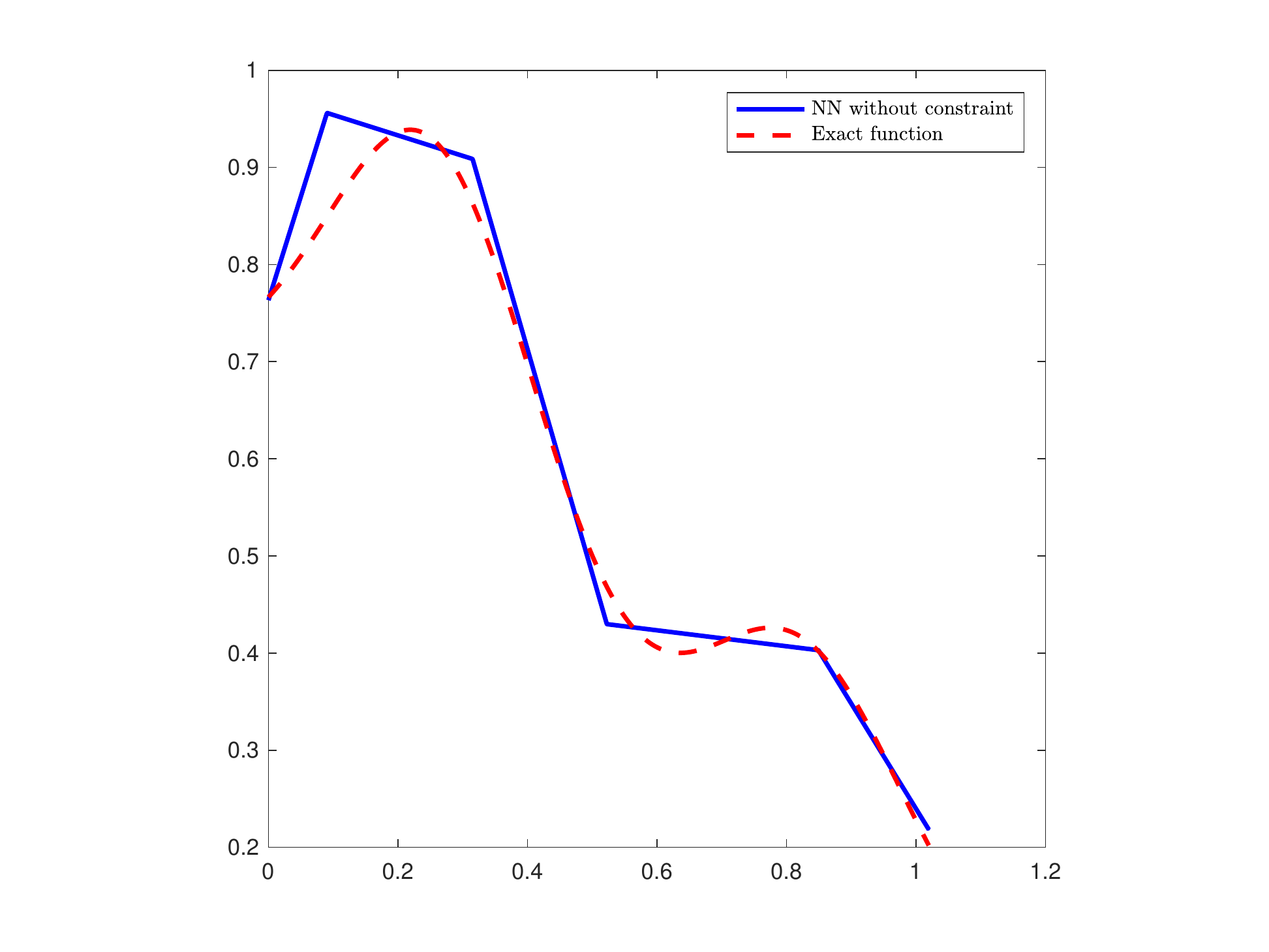} 
	\caption{Unconstrained: $y=0.2x$ cut}
  \end{subfigure} 
  \begin{subfigure}[b]{0.5\linewidth}
    \centering
	\includegraphics[width=1\linewidth]{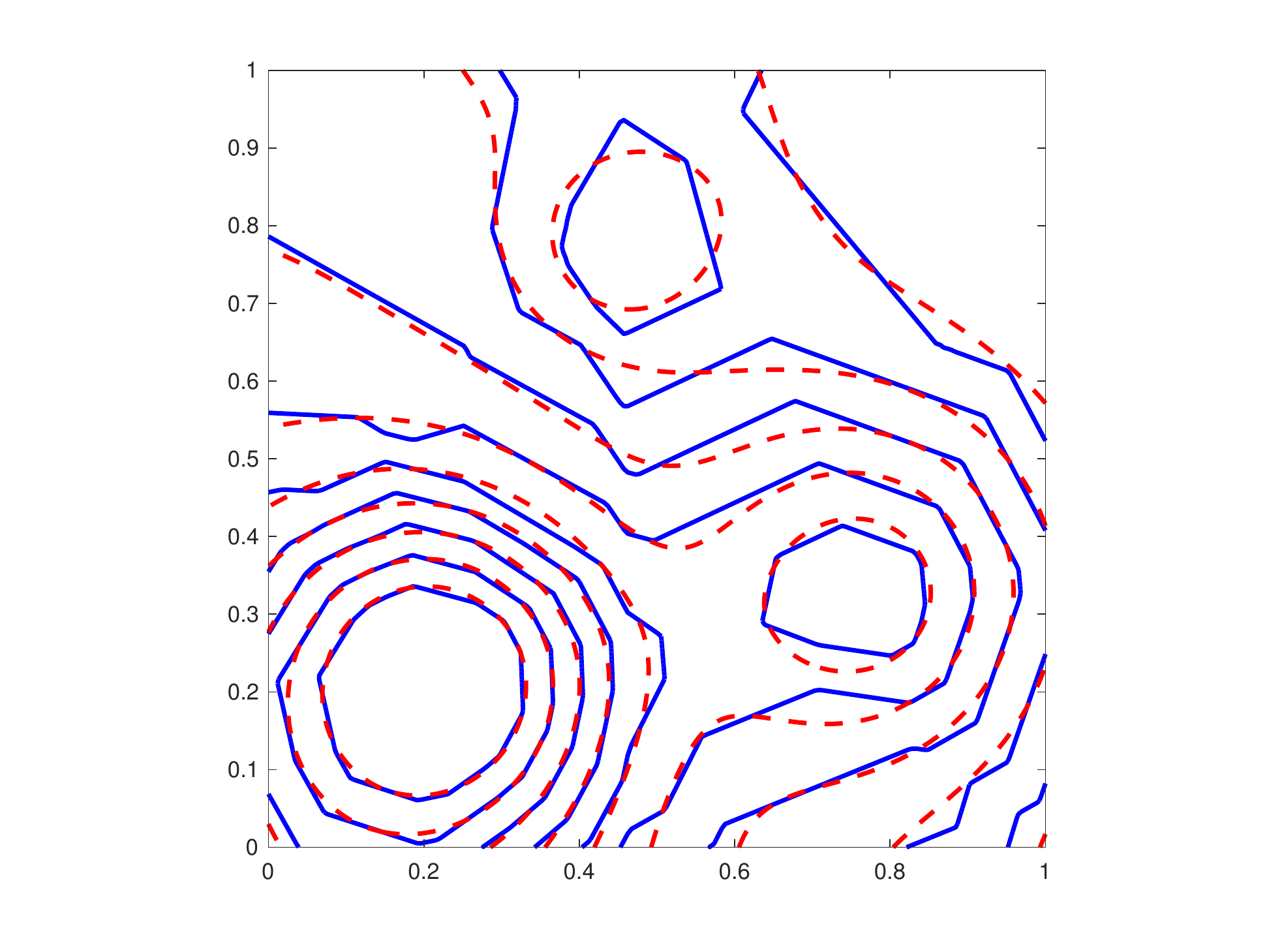} 
	\caption{Constrained: contours}
  \end{subfigure} 
  \begin{subfigure}[b]{0.5\linewidth}
    \centering
	\includegraphics[width=1\linewidth]{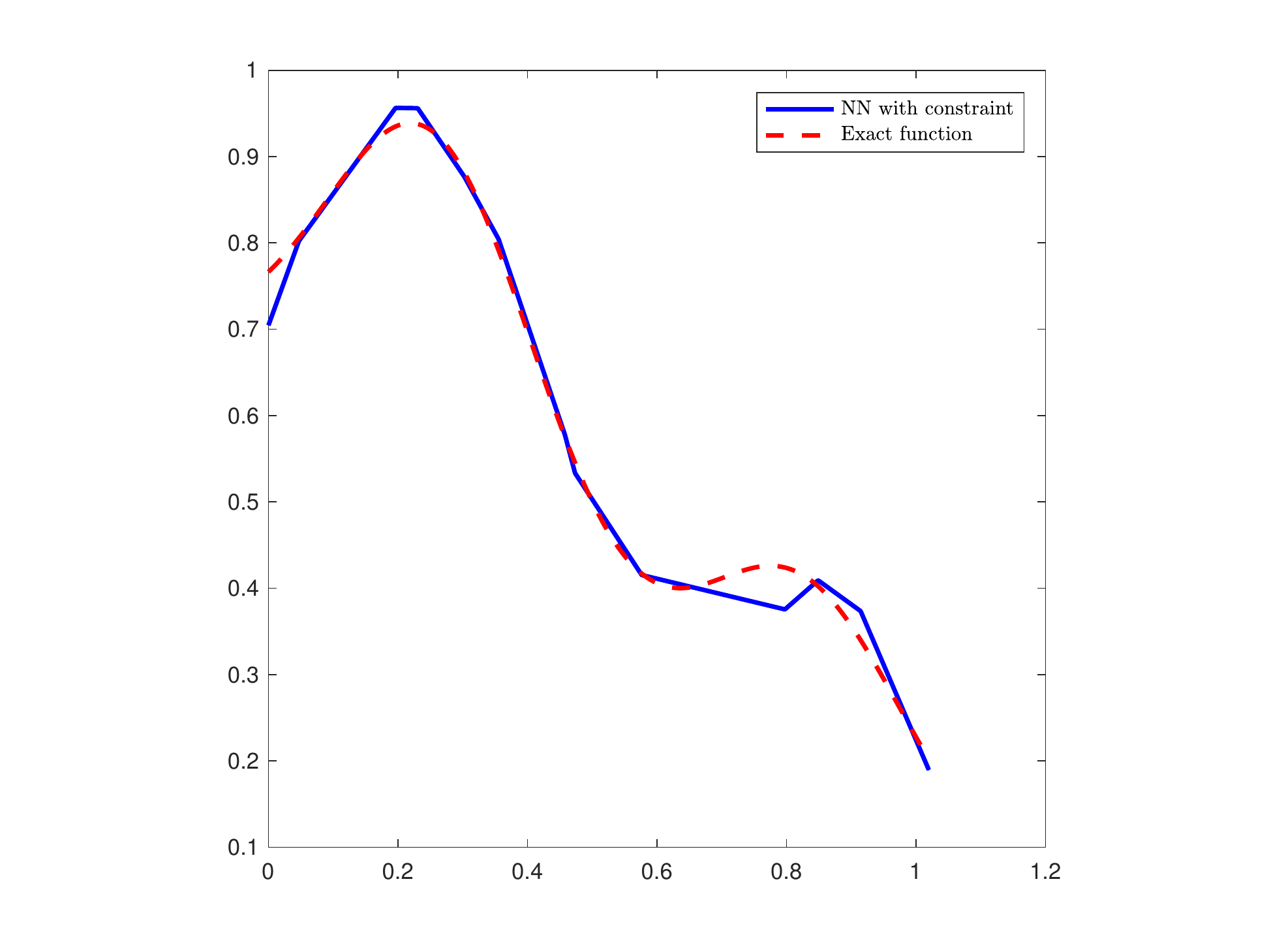} 
	\caption{Constrained: $y=0.2x$ cut}
  \end{subfigure}
  \caption{Numerical results 2D function \eqref{testf2} with NNs of
    the structure $\{2, 20, 10, 1\}$.
Top row: results by unconstrained NN formulation \eqref{MultNx};
Bottom row: results by constrained NN with \eqref{weight_general} and \eqref{b_general}.
Left column: contour plots; Right column: function cut along
$y=0.2x$. Dashed lines are the exact function.}
\label{fig:2L2D}
\end{figure}

\begin{figure}[ht] 
  \begin{subfigure}[b]{0.5\linewidth}
    \centering
	\includegraphics[width=1\linewidth]{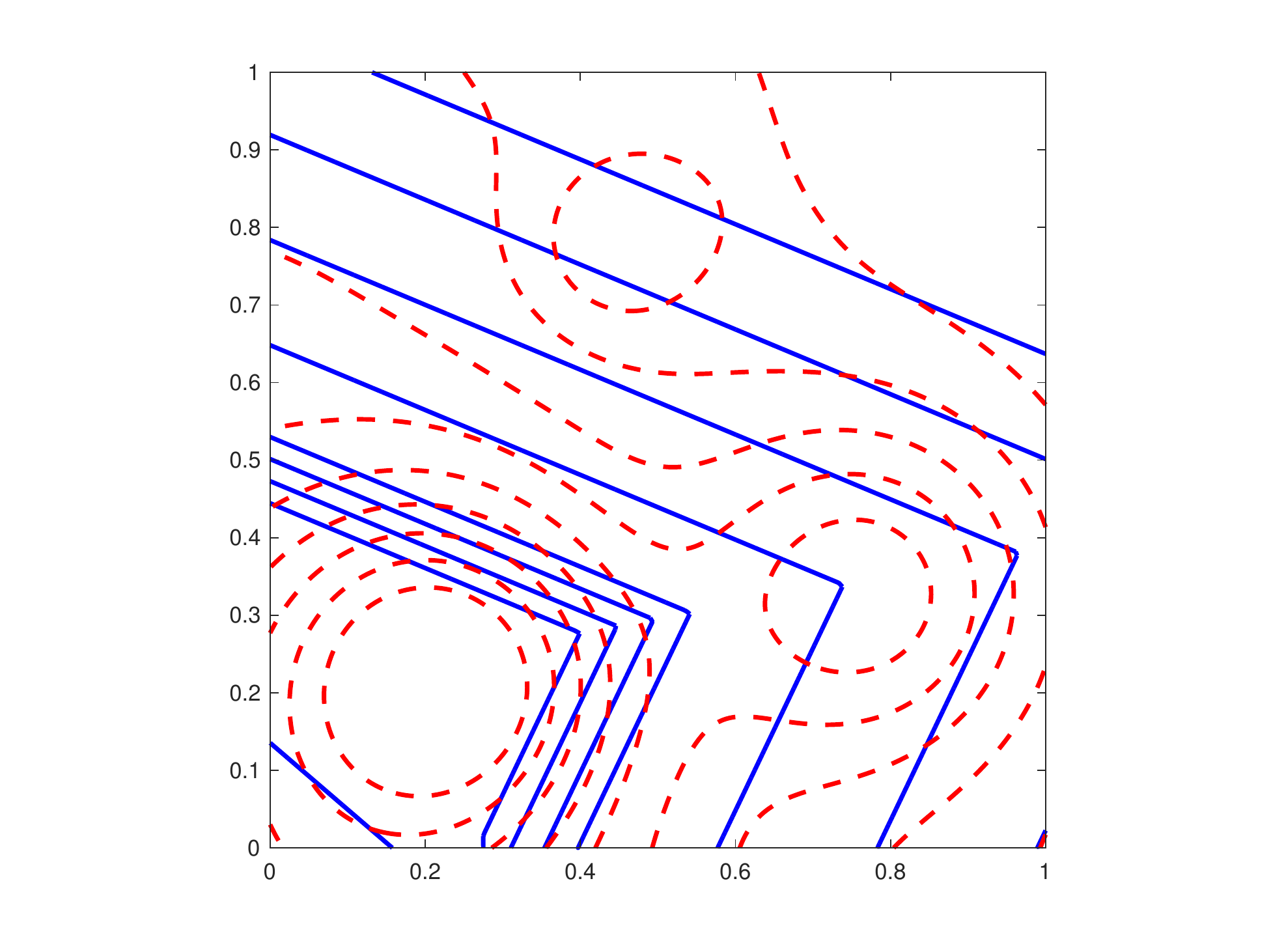} 
	\caption{Unconstrained: contours} 
  \end{subfigure}
	\begin{subfigure}[b]{0.5\linewidth}
    \centering
	\includegraphics[width=1\linewidth]{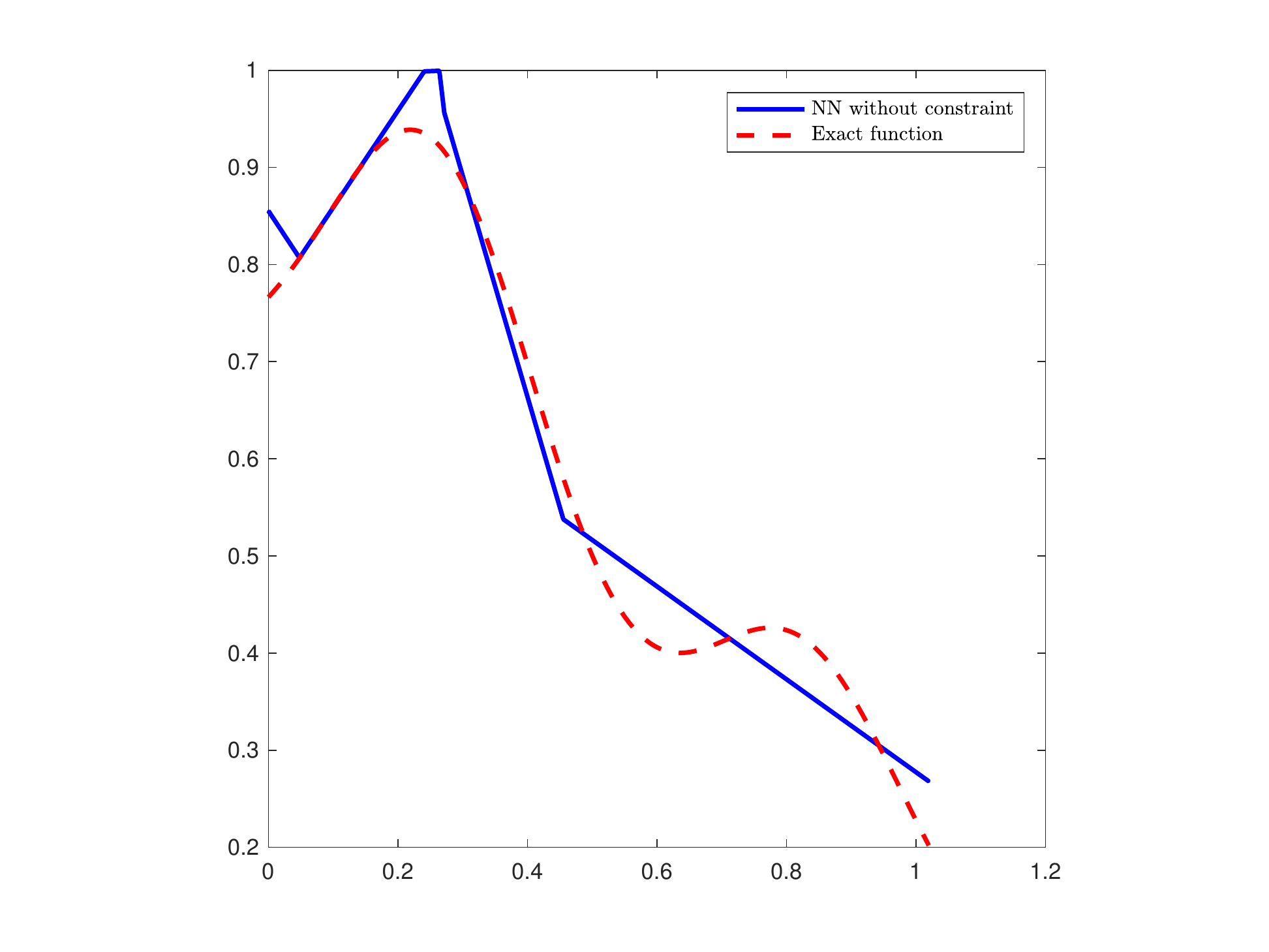} 
	\caption{Unconstrained: $y=0.2x$ cut} 
  \end{subfigure} 
  \begin{subfigure}[b]{0.5\linewidth}
    \centering
	\includegraphics[width=1\linewidth]{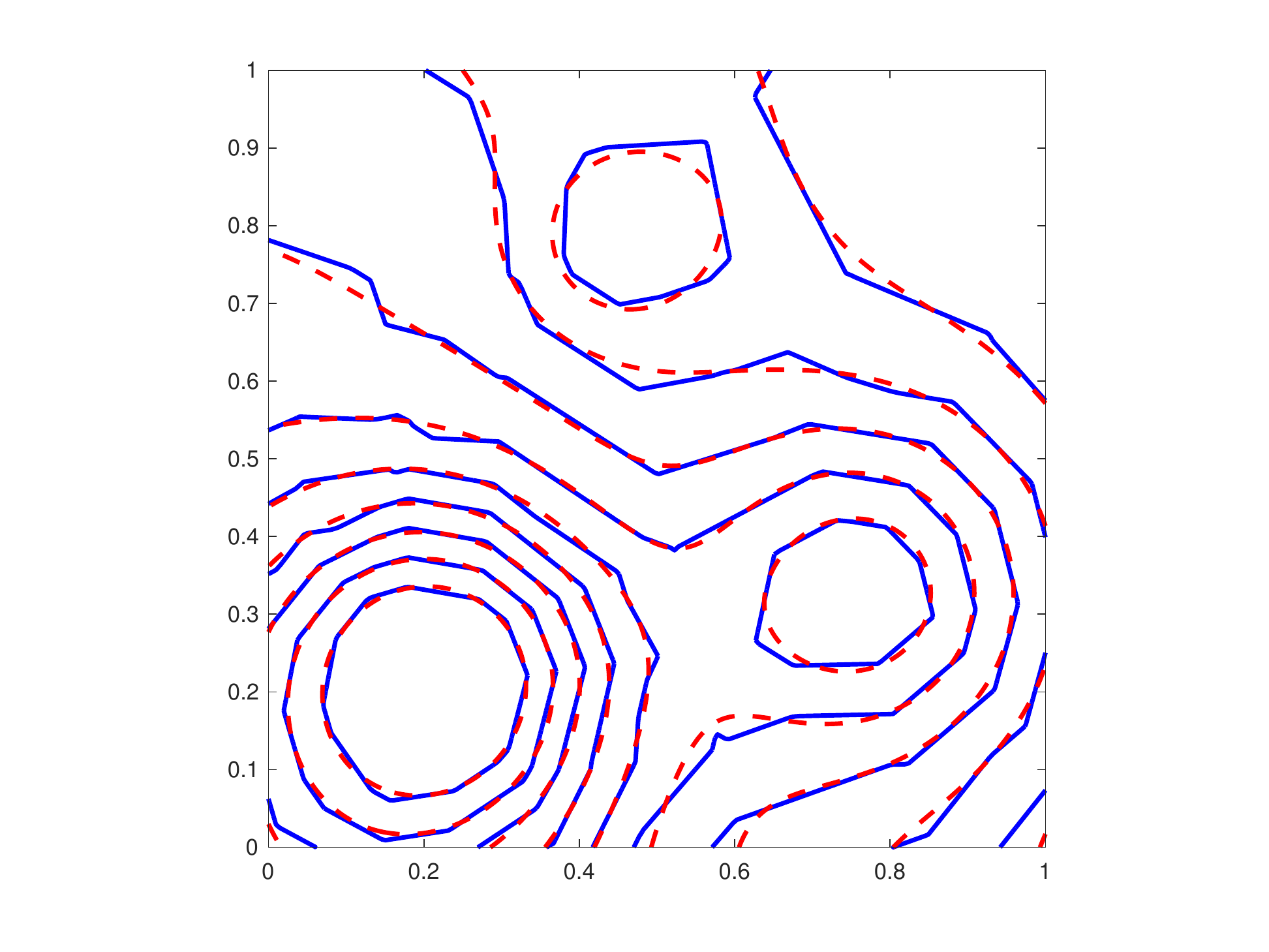} 
	\caption{Constrained: contours} 
  \end{subfigure} 
  \begin{subfigure}[b]{0.5\linewidth}
    \centering
	\includegraphics[width=1\linewidth]{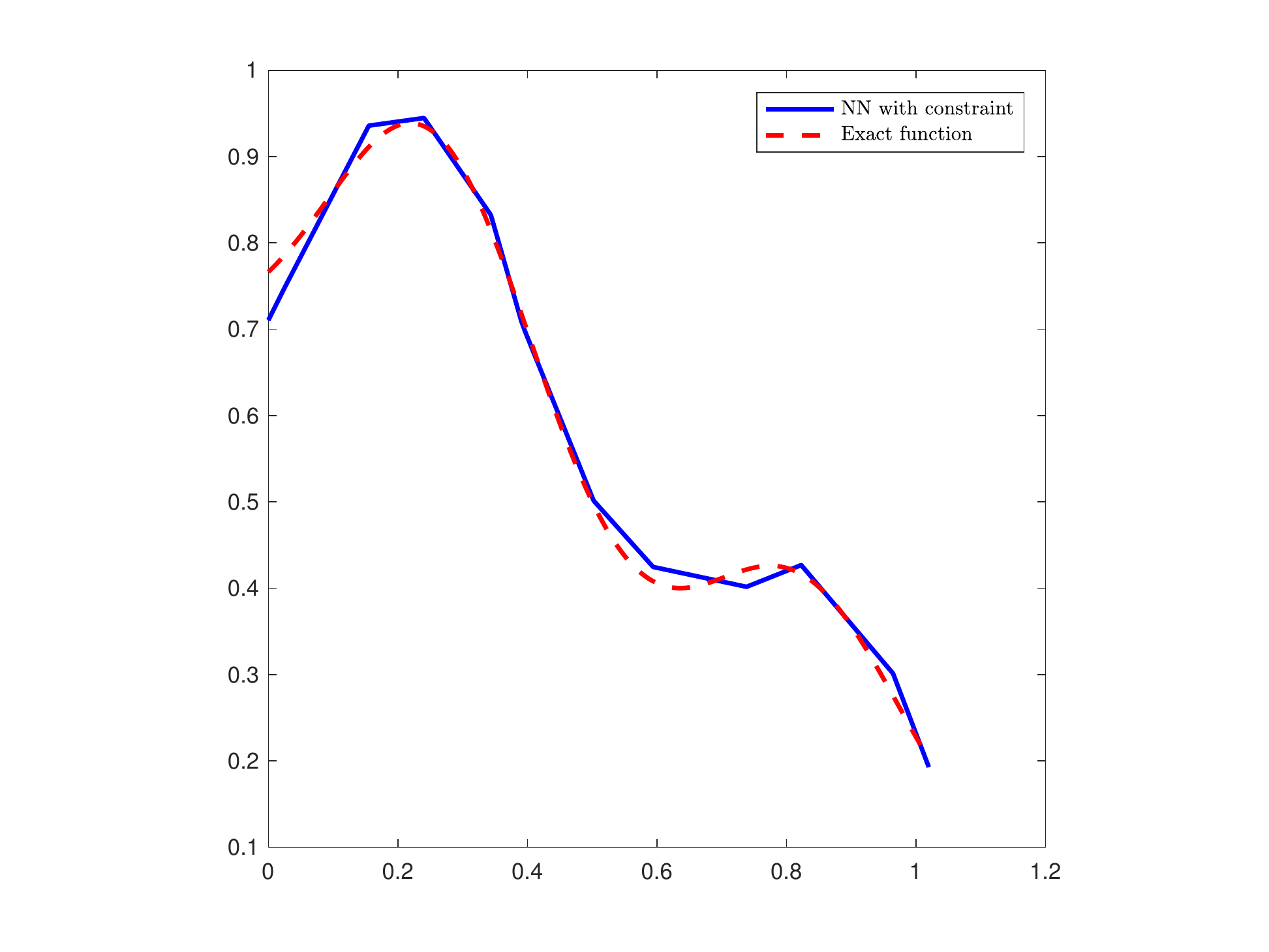} 
	\caption{Constrained: $y=0.2x$ cut}
  \end{subfigure}
  \caption{Numerical results 2D function \eqref{testf2} with NNs of
    the structure $\{2, 10, 10, 10, 10, 1\}$.
Top row: results by unconstrained NN formulation \eqref{MultNx};
Bottom row: results by constrained NN with \eqref{weight_general} and \eqref{b_general}.
Left column: contour plots; Right column: function cut along
$y=0.2x$. Dashed lines are the exact function.}
  \label{fig:4L2D}
\end{figure}

We now consider the two-dimensional Franke's function \eqref{testf2}. 
In Figure \ref{fig:2L2D}, the results by NNs with
$\{2, 20, 10, 1\}$ structure are shown. In Figure \ref{fig:4L2D}, the
results by NNs with  $\{2, 10, 10, 10, 10, 1\}$ structure are shown. 
Both the contour lines (with exactly the same contour values: from 0
to 1 with increment $0.1$) and the function value at $y=0.2x$ are
plotted, for both the unconstrained NN \eqref{MultNx} and the constrained
NN with the constraints \eqref{weight_general} and
\eqref{b_general}. Once again, the two cases use the same random
sequence for initialization. The results show again the notably
improvement of the training results by the constrained formulation.

\section{Summary} \label{sec:summary}

In this paper we presented a set of constraints on multi-layer
feedforward NNs with ReLU and binary activation functions. The weights
in each neuron are constrained on the unit sphere, as opposed to the
entire space. This effectively reduces the number of parameters in
weights by one per neuron. The threshold in each neuron is
constrained to a bounded interval, as opposed to the entire real
line. We prove that the constrained NN formulation is equivalent to
the standar unconstrained NN formulation. The constraints on the
parameters reduce the search space for network training and can
notably improve the training results. Our numerical examples for both
single hidden layer and multiple hidden layers verify this finding.



\clearpage
\bibliographystyle{plain}
\bibliography{neural}

\end{document}